%% file: icml_main.tex
\documentclass{article}

\usepackage{microtype}
\usepackage{graphicx}
\usepackage{subfigure}
\usepackage{booktabs} %

\usepackage{hyperref}

\usepackage[accepted]{icml2020}

\input{header.tex}

\icmltitlerunning{NSD Bandits with Intermediate Observations}

\begin{document}

\twocolumn[
\icmltitle{Non-Stationary Delayed Bandits with Intermediate Observations}

\icmlsetsymbol{equal}{*}

\begin{icmlauthorlist}
\icmlauthor{Claire Vernade}{equal,dm}
\icmlauthor{Andr\'as Gy\"orgy}{equal,dm}
\icmlauthor{Timothy A. Mann}{dm}
\end{icmlauthorlist}

\icmlaffiliation{dm}{DeepMind, London, UK}

\icmlcorrespondingauthor{Claire Vernade}{vernade@google.com}
\icmlcorrespondingauthor{Andr\'as Gy\"orgy}{agyorgy@google.com}

\icmlkeywords{Bandits, Non-Stationary, Delayed Feedback}

\vskip 0.3in
]

\printAffiliationsAndNotice{\icmlEqualContribution} %

\begin{abstract}
\vspace{-0.1cm}
Online recommender systems often face long delays in receiving feedback, especially when optimizing for some long-term metrics. While mitigating the effects of delays in learning is well-understood in stationary environments, the problem becomes much more challenging when the environment changes. In fact, if the timescale of the change is comparable to the delay, it is impossible to learn about the environment, since the available observations are already obsolete.
However, the arising issues can be addressed if intermediate signals are available without delay, such that given those signals, the long-term behavior of the system is stationary. To model this situation, we 
introduce the problem of stochastic, non-stationary, delayed bandits with intermediate observations. We develop a computationally efficient algorithm based on $\UCRL$, and prove sublinear regret guarantees for its performance. Experimental results demonstrate that our method is able to learn in non-stationary delayed environments where existing methods fail. 
\end{abstract}

\section{Introduction}

Optimizing long-term metrics is an important challenges for online recommender systems \citep{wu2017returning}. 
Consider, for instance, the media recommendation problem, where the final feedback on the content is only given after the customer spent time watching or reading it. 
Another major case is that of conversion optimization in online marketing \citep{yoshikawa2018nonparametric}: Here a user's click on an advertisement/recommendation might be observed almost immediately, but whether the click is converted into a long-term commitment, such as a purchase, can be observed usually only after some much longer delay \citep{chapelle2014modeling}, which causes modelling issues for robust inference.

Delayed feedback in online learning have been addressed both in the full information setting (see, e.g., \citealp{joulani2013online}, and the references therein), and in the bandit setting (see, e.g., \citealp{mandel2015queue,vernade2017stochastic, CBGM19}, and the references therein), assuming both stochastic and adversarial environments. The main take-away message from these studies is that, for bandits, the impact of a constant delay $D$ results in an extra additive $O(\sqrt{DT})$ term in the regret in adversarial settings, or an  additive $O(D)$ term in stochastic settings.
The aforementioned approaches often provide efficient and provably optimal algorithms for delayed-feedback scenarios. Nonetheless, the algorithms naturally  wait for having received  enough feedback  before actually learning something.
This is, in general, sufficient in stationary environments or when comparing with the best fixed action in the adversarial setting. Such methods, however, may be quite unsuitable in a non-stationary environment. 
For example, it is easy to see that in an environment that changes abruptly, every change results in an extra regret term mentioned above. An extreme example of a bad situation is when the environment changes (abruptly) every $D$ steps, in which case, by the time any feedback is received by the recommender system, its value becomes obsolete, thus no learning is possible. Accordingly, the extra regret terms above sum up to $T$, for example, in the stochastic case a regret of $D$ is suffered for each of the $T/D$ changes. 
Delays and non-stationarity are, in general, strongly entangled, which badly affects the performance of any online learning algorithm. 

In fact, real world problems are non-stationary in most cases. 
For example, trending or boredom phenomena may affect the conversion rates \citep{agarwal2009spatio}, forcing the systems to have estimators as local as possible in time.
As such, algorithms developed for non-stationary online learning (see, e.g., \citealp{auer2002finite,garivier2011upper,GyLiLu12,gajane2018sliding}) crucially depend on the availability of recent observations, which is often not possible in the presence of delays.
To alleviate this problem, practical systems often monitor -- and sometimes optimize for -- some proxy information instead of the real target. Typically, the web industry optimizes for click-through rate instead of conversions \citep{agarwal2009online}.

One step further in utilizing proxy information is to directly model the connection between the proxies and the final outcomes. This approach was formalized recently by \citet{mann2018learning}, who proposed a model where 
\emph{intermediate signals} can be observed without any delay, while the final feedback is independent of the system's action given the intermediate feedback, disentangling the effects of delays and non-stationarity. 
The model is based on the simple observation that even if the observation of the optimized metric is heavily delayed, %
the systems record many intermediate signals like clicks, labels or flags provided by the customers. In our media-recommendation example, this means that if a user decided to download a book, the probability of reading the book will be the same. While this model is also only a crude approximation of real-world scenarios, it allows to analyze the effect of such intermediate feedback signals in a principled way, and  \citet{mann2018learning} show both in theory and in practice that the proposed approach has significant benefits in full-information prediction problems under practical assumptions on the delay and the user population.

In this paper, we 
propose and analyze the bandit version of the model of \citet{mann2018learning}, which we call the stochastic non-stationary delayed bandit problem with intermediate observations (NSD for short). Here the system takes actions repeatedly, observes some intermediate feedback immediately, and the target metric after some (fixed) delay. It is assumed that the distribution of the intermediate feedback, called \emph{outcomes} or \emph{signals} in what follows, changes over time, while the final (delayed) metric, called the \emph{reward}, depends on the observed signal in a stationary way, and is conditionally independent of the action given the signals.
In this way, our model disentangles the effects of delays and non-stationarity, and hence algorithms leveraging intermediate outcomes can learn even in fast changing environment with large observation delays.
In the media-recommendation example, the actions of the systems are recommending books, which a user may or may not download, depending on the time-varying popularity of the recommended book. However, once the book is downloaded, we assume that the user will read it with a fixed (but unknown) probability.

We propose and analyze an algorithm for the NSD problem,
which can be thought of as a carefully derived variant of the $\UCRL$ algorithm of \citet{jaksch2010near} for our problem. Similarly to the sliding-window $\UCB$ ($\SWUCB$) algorithm for non-stationary stochastic bandits \cite{garivier2011upper} and the sliding-window $\UCRL$ ($\WUCRL$) algorithm of  \citet{gajane2018sliding}, our method uses sliding-window estimates for the non-stationary parameters of the problem, but these are combined with delayed estimates for the the stationary parameters.
We show both theoretically (through regret bounds) and in simulations that the proposed algorithm indeed disentangles the effects of the delay and non-stationarity.

The paper is organized as follows. The NSD bandit framework is presented in Section~\ref{sec:setting}. Our algorithm $\NSDUCRL$ is describe in Section~\ref{sec:algorithm}, while their performance is analyzed theoretically in Section~\ref{sec:analysis}. Experimental results are provided in   Section~\ref{sec:experiments}, while related work and future directions are discussed in Sections~\ref{sec:related} and~\ref{sec:conlusion}.


\section{Setting}
\label{sec:setting}

\paragraph{Notation. } For any integer $N>0$, let $\Delta_N \subset \R^N$ denote the simplex in $N$ dimensions, and let $N$ and $[N]:=\{1,\ldots N\}$. For any $x \in \R$, $(x)^+=\max \{0, x\}$ is the positive part of $x$. An MDP $M$ is characterized, in the tabular case, by $M=(\mathcal{A}, \mathcal{S}, P, \theta)$, i.e~a finite set of $S$ states $\mathcal{S}=[S]$ and $K$ actions $\mathcal{A}:=[K]$, a transition probability tensor $P$ and a matrix $\theta$ of reward parameters. Each element shall be precisely instantiated for our setting later.  

\paragraph{Learning scenario. }

We consider a stochastic bandit setting with a finite set of $K$ actions and a finite set of $S$ outcomes, with $K, S \ge 2$. The learning procedure at each round $t$ is the following:
\vspace{-0.3cm}
\begin{itemize}
    \item The learner chooses an action $A_t \in [K]$;\vspace{-0.15cm}
    \item A categorical  outcome $S_t\in [S]$ is revealed. It is assumed that $S_t$ is a categorical variable such that $S_t \sim p_t(a)$, where $p_t(a)\in \Delta_S$. This probability vector may change with time; \vspace{-0.15cm}
    \item After a possibly random delay $D_t$, a reward $R_t$ drawn from some fixed (but unknown) distribution that depends on the outcome $S_t$ is revealed to the learner. It is assumed that conditionally on $S_t$, the reward $R_t$ is independent of the action $A_t$, and it has mean $\theta_{S_t}$ (the vector of all means will be denoted by $\theta=(\theta_1,\ldots,\theta_S)^\top$). For simplicity, we assume throughout the paper that the rewards are $[0,1]$-valued and the delays are constant, $D_t=D$; our results can be extended, e.g., to random delays and sub-Gaussian reward distributions.
    \vspace{-0.15cm}
\end{itemize}

\begin{figure}
    \centering
    \includegraphics[width=0.9\columnwidth]{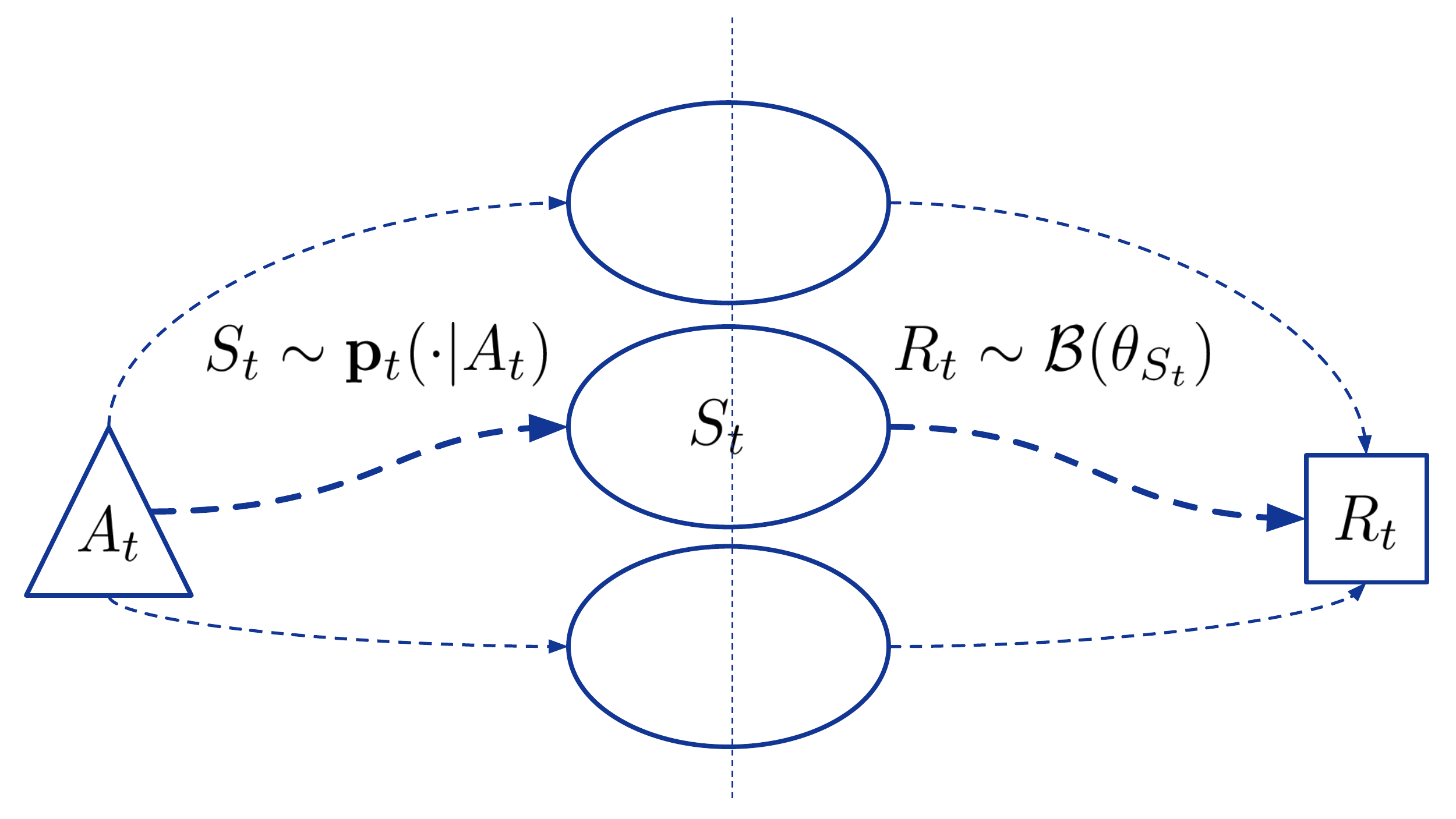}
    \vspace{-0.2cm}
    \caption{Schema of the MDP underlying the NSD bandit model with $L$ independent signals. Action $A_t\in[K]$ is taken in the triangle state (0) and a transition to a round signal state is observed. }
    \label{fig:mdp}
\end{figure}

Note that this factored bandit model can also be viewed as a simple episodic MDP $M=([S+1],[K],(P_{a,t})_{a\in[K]},\theta)$ with $S+1$ states and $K$ actions, as show in Figure~\ref{fig:mdp}. In state $0$, the learner takes an action $a\in [K]$ and transitions to a \emph{signal state}  $s \in [S]$ with probability $p_t(s|a)$ (we will use the notation $p_t(a)$ to denote the vector $(p_t(1|a),\ldots,p_t(S|a))^\top$ of transition probabilities). 
They fall back into state $0$ for the next round while a reward with mean $\theta_s$ is generated (but observed later).

\paragraph{Non-stationary transitions.}
We assume the environment is changing. There is a vast body of literature on non-stationary multi-armed bandits \cite{auer2002finite, garivier2011upper, besbes2014stochastic, auer2019achieving}, contextual bandits \cite{chen2019new} and reinforcement learning  \cite{jaksch2010near,gajane2018sliding} (tabular case).

In this work, we focus on switching environments: there exist $0\leq \Gamma_T <T$ instants where all transition probabilities abruptly change. At the same time, we assume that the parameter vector $\theta$ of the reward distributions remains fixed.
This models the scenario described in the introduction. The popularity of the items to be recommended changes with time, and hence the likelihood $p_t$ of the intermediate observations changes. However, given the intermediate outcomes, the reward distribution does not change. For instance, if a book is downloaded (this is the intermediate outcome) the probability that the user will read it is constant.
Note that we index $\Gamma_T$ with the horizon to emphasize that if the latter goes to infinity, the number of changes should do so as well, otherwise the impact of non-stationarity disappears in the asymptotic regime.  

\paragraph{Expected return and dynamic regret.}

For a given action $a \in [K]$, the expected reward at round $t$ is a function of the instantaneous parameters:
\[
\rho_t(a) = \sum_{s \in [S]} p_t(s |a)\theta_s = p_t(a)^\top \theta~.
\]
The optimal action at round $t$ is the one that has the highest expected reward under the current distribution over the outcomes:
\begin{equation}
    \label{eq:max-rew}
    \rho^*_t = \argmax_{a\in [K]} \sum_{s \in [S]} p_t(s|a) \theta_s = p_t(\cdot| a^*_t)^\top \theta ,
\end{equation}
where $a^*_t$ is the optimal action at round $t$, that is, the one achieving the maximum above. For simplicity, we assume it is unique. 

The goal of the learner is to minimize the dynamic expected regret defined as 
\begin{align*}
    \Reg(T) = \sum_{t=1}^T \rho^*_t - \rho_{t}(A_t).
\end{align*}
This metric is called \emph{dynamic} because the baseline is the learner that sequentially adapts to environment changes, which is a different and arguably harder problem than competing with the best action in hindsight. Because of our assumption that the environment changes only $\Gamma_T$ times, the optimal action $a^*_t$ also only changes at most $\Gamma_T$ times.

Note that the transition probabilities are not state-dependent as opposed to usual reinforcement learning (RL) settings because in our model there is only one state that allows to take actions. Following the action, a transition is observed to a state where no action can be taken, and only a future reward, assimilated to the state's expected value, is observed. One can think of our model as a disentanglement of the usual RL setting into a transition-then-reward bandit model.

\section{Algorithms}
\label{sec:algorithm}
In this section, we present a version of $\WUCRL$ of \citet{gajane2018sliding}, instantiated for our NSD bandit problem. 
Even though the general philosophy of the algorithm is fairly similar to the standard one, the simple structure of the MDP underlying our model does simplify things a bit. 

Our method, 
presented in Algorithm~\ref{alg:ucrl2}, relies on two types of confidence regions, for transition probabilities and rewards. 
Because of the non-stationarity of the transitions, we estimate them using a sliding window with a fixed window size $W>0$, given as input to the learner. 
In contrast, the reward parameters are fixed so no window is needed to estimate them. 
At every round $1\leq t\leq T$, for each action $a\in [K]$, we define the windowed counts
\[
N^W_t(a) = %
\max\left\{1,\sum_{u=(t-W)^+}^{t-1} \ind\{A_u =a\}\right\},
\]
and for all $s\in [S]$, we estimate the transition probability as
\begin{equation}
\label{eq:trans_prob}
\hat{p}_t(s|a) =  \frac{\sum_{u=(t-W)^+}^{t-1}\ind\{A_u=a,S_u=s\}}{N^{W}_t(a)}~.
\end{equation}

On the other hand, for each signal $s\in [S]$, we keep the usual counts $N^D_t(s)= \max\left\{1,\sum_{u<t-D} \ind\{S_u =s\}\right\}$ (recall that observing rewards are delayed by $D$, so at the beginning of round $t$ only rewards for rounds before $t-D$ are available), and the empirical average estimator 
\[
\hat{\theta}_t(s)=\frac{1}{N^D_t(s)}\sum_{u<t-D} R_t \ind\{S_u =s\}
\]
of the rewards up to round $t-D$. %
As in standard stochastic bandit models, 
we consider high-probability upper-confidence bounds for some fixed $\delta>0$:
\begin{equation}
\label{eq:U_t}
   U_t(s) =  \min\left\{1, \hat{\theta}_t(s)+\sqrt{\frac{C_{T,\delta}}{N^D_t(s)}} \right\}.
\end{equation}
where $C_{T,\delta}=2\log(2TS/\delta)$.
The key step in the learning is to search for the most optimistic MDP in a high-probability parameter space: for some $\delta>0$ we define $C_{W,T,\delta}=2S\log(KWT/\delta)$, and for each action $a\in[K]$,
\begin{align}
\label{eq:rho}
\!\!\!\rho^+_t\!(a)
&= \max_{q \in \Delta_S} \left\{q^\top U_t\!: \|\hat{p}_t(a) -q \|_1 \leq \!\sqrt{\frac{C_{W,T,\delta}}{N^W_t(a)}}\,\right\}\!. %
\end{align}
Here $q$ ranges over the plausible candidate set of transition probabilities for $p_t(a)$ with estimated average reward $q^\top U_t = \sum_{s\in[S]} q(s)U_t(s)$. Note that since $U_t(s)$ is an optimistic (upper) estimate of $\theta_s$ (with high probability), $\rho^+_t(a)$ is an optimistic estimate of $\rho_t(a)$. We denote the value of $q$ achieving the maximum above by $\tp_t(a)$.
The algorithm then acts optimistically by choosing the action with the largest $\rho^+_t$ index. 

This mechanism is reminiscent to the Extended Value Iteration algorithm of \cite{jaksch2010near}, but we only need a simplified version, which is given in Appendix~\ref{ap:evi}. 
Further justifications on the choices of confidence regions are provided in the next section where they are used to prove high-probability upper bounds on the regret. 
Note that as opposed to the general $\UCRL$ algorithm of \cite{jaksch2010near}, our method does not require phases. 

\begin{algorithm}%
\caption{$\NSDUCRL$ for NSD-Bandits  \label{alg:ucrl2}}
\begin{algorithmic}%
\STATE{\textbf{Input:} $K$: number of arms, $S$: number of outcomes, $W$: window size, $\delta$: error probability}%
\STATE{ \textbf{Initialization}:  $\forall \, a\in [K]$ and  $\forall \, s \in [S]$:
\STATE Global counts:  $N_0(a)= N_0(a,s)=0$,  $N^D_0(s)=0$.}
\STATE{Local counts:  $N^W_0(a)=0$, $\hat{p}_0(\cdot|a)= \frac{1}{S}\ind_S$}
\STATE{Pull each action once.}
\FOR{$t \geq K+1$}
    \STATE  $\forall \, a\in [K]$, compute $\hat{p}_t(\cdot|a)$ as in \eqref{eq:trans_prob};
    \STATE $\forall \, s\in [S]$, compute $\hat{\theta}_t(s)$ and the high-probability upper bounds $U_t(s)$ as in \eqref{eq:U_t}.
    \STATE $\forall \, a\in [K]$, compute $\rho^+_t(a)$ according to \eqref{eq:rho} (by Algorithm~\ref{alg:EVI} in \cref{ap:evi}).
    \STATE Pull action $A_t \in \argmax_{a \in [K]} \rho^+(a)$;
    \STATE{Action update: $N^{W}_t(a)$ for all $a\in[K]$;}
    \STATE Signal update: 
    Receive the intermediate signal $S_t$ (with no delay) and update the windowed transition counts;
    \STATE Receive feedback $R_{t-D}$ for round $t-D$ and signal $S_{t-D}$.
\ENDFOR
\end{algorithmic}
\end{algorithm}

\section{Analysis}
\label{sec:analysis}

In this section we present high-probability upper bounds on %
the regret of $\NSDUCRL$. 

\subsection{Warm-up: Analysis in the stationary case}

As a first step, we provide an analysis of $\NSDUCRL$ when the transition probabilities are stationary, that is, $\Gamma_T=0$. 
This initial result gives an insight on how much regret is suffered due to the use of a sliding-window estimator.

\begin{theorem}
\label{th:stationary_regret}
Let $M=(S,K,P,\theta)$ define a stationary NSD problem with constant delay $D>0$. With probability at least $1-2\delta$, the regret of $\NSDUCRL$ with window size $W$ is bounded from above by  
\begin{align*}
\lefteqn{\Reg(T)  \leq O \left( T\sqrt{\frac{KS\log\left(\frac{KWT}{\delta}\right)}{W}} \right.} \\
& + \left.\left(\sqrt{S}\!+\!1\!+\!\sqrt{\log(1/\delta)}\right)\sqrt{T\log\left(\frac{TS}{\delta}\right)} + S(D\!+\!1) \right).
\end{align*}

With the choice $W=\Theta(T)$, we get
\begin{align*}
\lefteqn{\Reg(t) \leq O\left(\sqrt{TKS\log\left(\frac{KT}{\delta}\right)} \right.} \\ 
&\left. + \left(\sqrt{S}\!+\!1\!+\!\sqrt{\log(1/\delta)}\right)\sqrt{T\log\left(\frac{TS}{\delta}\right)} +S(D\!+\!1)\right).
\end{align*}

\end{theorem}
The proof of the theorem is deferred to Appendix~\ref{ap:stationary_regret}. 
It is based on standard techniques used in the literature to derive regret bounds for optimistic algorithms for bandit and MDP problems (mostly on the proof of the minimax bound of \UCRL). The non-standard parts in our derivation come from the fact that the transitions are estimated only based on the most recent window, but without delay, while the rewards can be estimated based on all observations, but these are delayed. Therefore, we need to disentangle the estimation of the transitions and rewards.

\begin{remark}[Random delays] \em
It is possible to extend our analysis used in the proof of Theorem~\ref{th:stationary_regret} for random delays. When the delays are random, the number of missing observations takes the role of $D$. The only part of the derivation affected by this change (see the proof for details) is the term $(D_t(s)+1)/(2(N_t(s)-D_t(s)-1)^{3/2})$, where now $D_t(s)$ is the number of missing observations from signal $s$ at time $t$. Under mild assumptions on the delays, $D_t(s) \le N_t(s)/2$ for $t>T_0$ with high probability where $T_0$ is large enough. Then the effect of delays, in expectation, is $O(T_0 + \E[D(s)])$, where $\E[D(s)] \approx \E[D_t(s)]$ for $t>T_0$. 
\end{remark}

\subsection{Main Results for NSD Bandits}

We can now prove the main results of this section, high-probability upper bounds on the regret of $\NSDUCRL$ in non-stationary and delayed environments. We start with a problem-independent bound, which is the extension of Theorem~\ref{th:stationary_regret} to the non-stationary case.

\begin{theorem}
\label{th:main_regret}
Let $M_T=(S,K,(P_t)_{t\leq T},\theta)$ define an NSD problem with $\Gamma_T$ switches and constant delays $D>0$. With probability at least $1-2\delta$,  the regret of $\NSDUCRL$ is bounded from above by
\vspace{-0.2cm}
\begin{align*}
\lefteqn{\Reg(T) \leq O\Bigg( W\Gamma_T + 2T\sqrt{\frac{KS\log\left(\frac{KWT}{\delta}\right)}{W}}} \\[-0.8em]
&+ \left(\sqrt{S}\!+\!1\!+\!\sqrt{\log(1/\delta)}\right)\sqrt{T\log\left(\frac{TS}{\delta}\right)} +S(D\!+\!1)\Bigg).
\end{align*}

Choosing 
$W=(T/\Gamma_T)^{2/3}(KS \log(KT/\delta))^{1/3}$,
we get 
\vspace{-0.3cm}
\begin{multline*}
\lefteqn{\Reg (T) \leq O\Big(T^{2/3} \left(\Gamma_T KS\log(KT/\delta)\right)^{1/3} } \\
+ S(D+1) + \sqrt{T(S+\log(1/\delta))\log(TS/\delta)}\Big).
\end{multline*}
\end{theorem}
\vspace{-0.5cm}
The proof of the theorem -- presented in Appendix~\ref{ap:main_proof} -- relies on the regret guarantees of $\NSDUCRL$ proved in Theorem~\ref{th:stationary_regret}. In short, after each switch, the estimators are all wrong for $W$ rounds, resulting in a linear $O(W)$ regret per switch. In between those phases, the algorithm adapts and suffers the sublinear regret proved in Theorem~\ref{th:stationary_regret}.  The subtlety of the analysis lies in the separate control of the reward estimation that does not suffer from switches.

Next we present a bound on the number of rounds where suboptimal actions are selected, as well as a problem dependent regret bound. We start with a few more definitions.
Let $\mathcal{T} (W) = \{1\leq t\leq T:\, \forall k\in [K], \, \forall t-W\leq s \leq t,\, p_s(k)= p_t(k) \}$ denote the round indices when all arms have had stable transition probabilities for at least $W$ rounds. An action $a \in [K]$ is called $\eps$-bad in time step $t$, if its gap is at least $\eps$, that is, if $\Delta_{t,a}=p_t(a^*_t)^\top \theta - p_t(a)^\top \theta \ge \eps$, and let $\ebad_t$ denote the set of $\eps$-bad actions in round $t$. The minimum $\eps$-bad observation probability for each $s \in [S]$ is defined as
$\pmse=\min\{p(s|a): t \in [T], a \in \ebad_t, p(s|a)>0\}$, and let $p_{\min}=\min_{s \in [S]} p^0_{\min}(s)=\min\{p(s|a): t \in [T], a \neq a^*_t, s \in [S], p(s|a)>0\}$. Finally, let $\Delta_{\min}=\min_{a,t: \Delta_{t,a}>0} \Delta_{t,a}$ denote the minimum gap of the actions over the time horizon.

\begin{theorem}
\label{th:problem-dependent}
Under the assumptions of Theorem~\ref{th:main_regret}, with probability at least $1-3\delta$, the number of times $\NSDUCRL$ selects $\eps$-bad actions in $\cT(W)$ is bounded by
\begin{align*}
O\!\left(\frac{T}{W} \frac{SK \log(\frac{KWT}{\delta})}{\eps^2} \!+\! SD\! +\! \sum_{s \in [S]}
\frac{\log(\frac{e}{\pmse})}{\pmse} \frac{\log(\frac{TS}{\delta})}{\eps^2} \right)\!. 
\end{align*}
Furthermore, the regret of $\NSDUCRL$ can be bounded by
\begin{align*}
O\Bigg(\frac{S}{\Delta_{\min}}
\Bigg(\frac{T\log(\frac{KWT}{\delta})}{W} &+ \frac{\log(\frac{e}{p_{\min}})\log(\frac{TS}{\delta})}{p_{\min}} \Bigg)  \\
& \qquad\qquad + W \Gamma_T + SD  \Bigg)\!. \\
\end{align*}
\end{theorem}
The proof is deferred to Appendix~\ref{ap:prob_dep_bounds}. It is based on a decomposition of the suboptimal actions taken into two groups depending whether the rewards of the actually reachable intermediate signals are estimated up to a required accuracy. 

\paragraph{Lower bound for standard, signal-agnostic methods. }
Following the construction of \citet{garivier2011upper}, it is easy to show a lower bound for signal-agnostic algorithms (i.e., algorithms that do not use the intermediate signals). 
Indeed, if the reward distributions can be selected arbitrarily for every stationary segment, the regret of a signal-agnostic algorithm is bounded from below by the regret of the same algorithm receiving the information of the change points and restarting (optimally) after each of them. Then, on every stationary segment, the minimax regret lower bound for the $K$-armed bandit problem applies (see, e.g., Chapter~15 of \citealp{lattimore2020bandit}), which, together with the effect of the delay $D$ on this segment, gives a lower bound of $\Omega(\sqrt{K l}+D)$ regret for a segment of length $l$. Considering environments with stationary segments of equal length $l=T/\Gamma_T$ term, gives the following lower bound:\footnote{This argument can be made precise by carefully considering that the lengths of the segments are integers and selecting independently the reward distribution for each segment from the minimax lower bound construction.}

\begin{proposition}
\label{pr:lower}
For any $\Gamma_T<T$ and any signal-agnostic algorithm, there exists an NSD problem such that the regret is lower bounded as
\[
\Reg(T) \geq \Omega \left( \min\left\{T,\sqrt{K (\Gamma_T+1) T} + (\Gamma_T+1) D\right\}\right).
\]
\end{proposition}

\subsection{Discussion}
Proposition~\ref{pr:lower} shows that for bandit algorithms, delays and non-stationarity are entangled and have a combined impact on minimax regret bounds. In contrast, by leveraging intermediate outcomes, $\NSDUCRL$ \emph{disentangles} this effect and only pays the price of the delay once, and not every time a new stationary segment starts (see Theorems~\ref{th:main_regret} and ~\ref{th:problem-dependent}).
This allows $\NSDUCRL$ to achieve sublinear regret in situations where standard, signal-agnostic algorithms would suffer linear regret. For example, if $D = \Omega(T/(\Gamma_T+1))$, the latter suffer linear regret (as the lower bound becomes $\Omega(T)$), while the bounds for our method guarantee sublinear regret as long as $S$ is small enough relative to $\Gamma_T$ (neglecting, of course, other conditions, like $KS\Gamma_T=o(T)$).
This emphasizes the main advantage of our feedback model: by disentangling delay and non-stationarity, learning is possible in a much broader family of situations.

Our two upper bounds provide slightly different insights to our algorithm:
Theorem~\ref{th:main_regret} shows an $\tilde{O}(\Gamma_T^{1/3}T^{2/3}+D)$ upper bound on the regret. Theorem~\ref{th:problem-dependent} provides a problem dependent bound that is essentially $\tilde{O}(\frac{T\log T}{W \Delta_{\min}}+ W \Gamma_T + D)$, which depends on the minimum gap of all the actions over the time horizon, and also on the minimum observation probability of the intermediate signals. It also shows that our method cannot suffer large losses too often, as it chooses $\eps$-bad actions at most $\tilde{O}(\frac{T}{W}\frac{\log T}{\eps^2}+W \Gamma_T + D)$ times. 

In case the problem is stationary, setting the window size to $W=T$ essentially recovers the best possible rates in both theorems, including the additive effect of the delay \citep{joulani2013online}. The $T^{2/3}$ regret rate of Theorem~\ref{th:main_regret} also agrees with the rates derived for MDPs in non-stationary environments \citep{jaksch2010near,gajane2018sliding}. The regret guarantee of Theorem~\ref{th:problem-dependent} improves upon a similar bound of \citet{garivier2011upper} for the non-stationary bandit problem (without delay), where the dependence on $\Delta_{\min}$ is quadratic. Similarly to them, we can also obtain a problem-dependent regret bound of $\tilde{O}(\sqrt{\Gamma_T T/\Delta_{\min}} + D)$, which requires setting the window size as $W=\Theta(\sqrt{T/\Delta_{\min}})$. The bound becomes a bit worse in $\Delta_{\min}$ if $W$ is tuned independently of this quantity. The main message of Theorem~\ref{th:problem-dependent}, however, is that if the gaps are not too small, the algorithm can learn really fast, while handling partially delayed observations.

 On the other hand, a $\tilde{O}(\sqrt{\Gamma_T T})$ minimax regret is possible for non-stationary bandits, and such a result automatically extends to shortest path problems with fix horizon, an instance of which we consider here. Nevertheless, to our knowledge, no practical algorithm exists for MDPs in general that achieves the minimax rate in non-stationary, abruptly changing environments considered in this paper. The task is also not made easier by the combined effects of delays and non-stationarity in our specific problem, which need to be disentangled to utilize the full power of our model. This comes with its own limitations; for example, we could not built on the \UBEV algorithm of \cite{DannLB17} (designed for the stochastic shortest part problem), which improves upon the guarantees of \UCRL by jointly estimating the transition probabilities and the rewards (value function in their case).

Getting our $\tilde{O}(\Gamma_T^{1/3}T^{2/3}+D)$ or $\tilde{O}(\sqrt{\Gamma_T T}/\Delta_{\min} + D)$ regret rates requires setting the window parameter $W$ using some prior information on the number of switches and on the horizon. This is common practice in the literature (see, e.g.,  \citealp{auer2002nonstochastic,garivier2011upper,besbes2014stochastic}, as
in face of non-stationarity, it is challenging to design fully adaptive online learning algorithms that do not rely on restricting the length of the history they use. Indeed, in absence of other assumptions on the underlying non-stationary function controlling the rewards, the best a learner can do is to  forget the past to maintain estimators as little biased as possible. 
The optimal behavior should then be to forget the past \emph{adaptively}. 

This is the recent approach taken by \cite{auer2019achieving,chen2019new} for the standard and the contextual multi-armed bandit problems, respectively. They include careful change-detection procedures in their algorithms to detect (almost) stationary segments, and obtain the first fully adaptive $O(\sqrt{\Gamma_T T})$ bounds on the dynamic regret with $\Gamma_t$ switches.
It seems straightforward to show that, in a delayed environment, the regret of these methods matches (up to logarithmic factors) the rate of the lower bound of Proposition~\ref{pr:lower}.

These algorithms pull all arms alternately until they identify the current best one and then commit to it and keep a carefully tuned change-point safety check in parallel. 
It would be interesting to adapt these methods for learning the transition probabilities in our algorithm in order to exploit the fully adaptive ideas mentioned above. 
We conjecture that the optimal regret for NSD bandits with intermediate observations should be bounded by $\tilde{O}(\sqrt{(\Gamma_T+1) K S T}+D)$, but we leave this question as an open problem for future work on this topic.

In all our regret bounds, we treat $S$ as a constant term. Comparing our regret bounds (Theorems~\ref{th:main_regret} and~\ref{th:problem-dependent}) to the rates achievable by the standard signal-agnostic methods (Proposition~\ref{pr:lower}) shows that $\NSDUCRL$ behaves better in terms of the delay than signal-agnostic algorithms when $\Gamma_T$ is larger than $S$. 
However, a large value of $S$ may deteriorate the first, time-dependent term in our bounds, possibly making $\NSDUCRL$ theoretically inferior to signal-agnostic methods in such cases.
Deciding whether this is just an artifact of our analysis or a real phenomenon is left for future work. In particular, it could be interesting to derive a problem-dependent lower bound for this new family of structured bandit problems, similarly to those of \citet{graves1997asymptotically} for the standard setting.

\section{Experiments}
\label{sec:experiments}

In this section we present experimental results to provide more insight on the impact and potential usefulness of the intermediate signals for different regimes of delays with respect to $\Gamma_T /T$.
We test our algorithm in different scenarios by comparing it to standard bandit algorithms that are agnostic to the intermediate signals, as well as to oracle bandit strategies that have access to various extra information. 

\paragraph{Baselines.}
First, natural but weak baselines are the standard $\UCB$ algorithm \cite{auer2002finite} and its sliding-window version $\SWUCB$ \cite{garivier2011upper}, which ignore the intermediate observations. 
Those policies have sublinear regret in stationary and non-stationary environments respectively. 
Thus, they should provide reasonable performance in our problem (when the delays are not too large), but are expected to be inferior compared to our method, proving that signals at the very least do not hurt. 
As a Bayesian alternative to $\NSDUCRL$, we implemented $\NSDPSRL$, based on the posterior sampling reinforcement learning ($\PSRL$) algorithm of \citet{osband2017posterior} (details are given in Appendix~\ref{ap:psrl}). We expect this method to have similar performance to $\NSDUCRL$.
We also construct two stronger oracle policies. $\OrUCB$ is a signal-agnostic UCB policy that receives the extra-information of the change-points and restarts optimally after each of them. $\OrNSD$ is $\NSDUCRL$ without windowing but with oracle restarts as well. We also consider full oracles that do not suffer delays, and call them respectively $\OrNSDnd$ and $\OrUCBnd$. Restarting is the optimal behavior in a non-stationary environment so these oracles should all outperform any other algorithms in this setting.

\paragraph{Experimental setup.} Throughout the experiments we use the transition probabilities and mean rewards as reported in Table~\ref{tbl:experiment}.
\begin{table}
    \centering
    \begin{tabular}{c|c|c|c|| c}
    reward & $\theta_1$ & $\theta_2$ & $\theta_3$ & \\ \hline
           & 0.8 & 0.4 & 0.2 & \\\hline \hline
       action ($a$)  & $p(1|a)$ & $p(2|a)$ & $p(3|a)$ & $\rho_a$ \\ \hline
         1 & 0.8 & 0.1 & 0.1 &  0.7 \\  \hline
         2 & 0.1 & 0.8 & 0.1 & 0.42 \\ \hline
         3 & 0.8 & 0.1 & 0.8 &  0.28\\ \hline
         4 & 0.1 & 0.4 & 0.5 & 0.34 
    \end{tabular}
\caption{Experiment setting}
\vspace{-0.5cm}
        \label{tbl:experiment}
    \label{tab:exp_setting}
\end{table}
The delay parameter $D$ and the window size $W$ of the algorithm are discussed in the dedicated sections. 
To simulate non-stationarity, we draw a random integer $r\in \{1,\ldots,K-1\}$ at each change point and we permute the action vectors of the transition matrix by shifting them $r$ times. So arm $i$ becomes arm $(i+r-1 \mod K)+1$. This has the advantage of ensuring the best arm always changes after a change point. 
Note that a side effect of this construction is that after a few changes, it might be the case that an arm that had better values on average so far suddenly becomes the best one, which allows agnostic policies like $\UCB$ to perform well on such a phase (typically the last one in our experiments). But this is an artifact of this experimental setup and we tried our best to design experiments that do not suffer from it. 

Unless otherwise stated, we run experiments with time horizon $T=8000$ with $\Gamma_T=3$ change points at rounds $\{2000,4000,6000\}$. All results presented are averaged over $50$ independent runs. The shaded areas in the graphs correspond to the 95\% confidence regions.

\begin{figure}[t]
\vspace{-0.1cm}
    \centering
    \includegraphics[width=0.8\columnwidth]{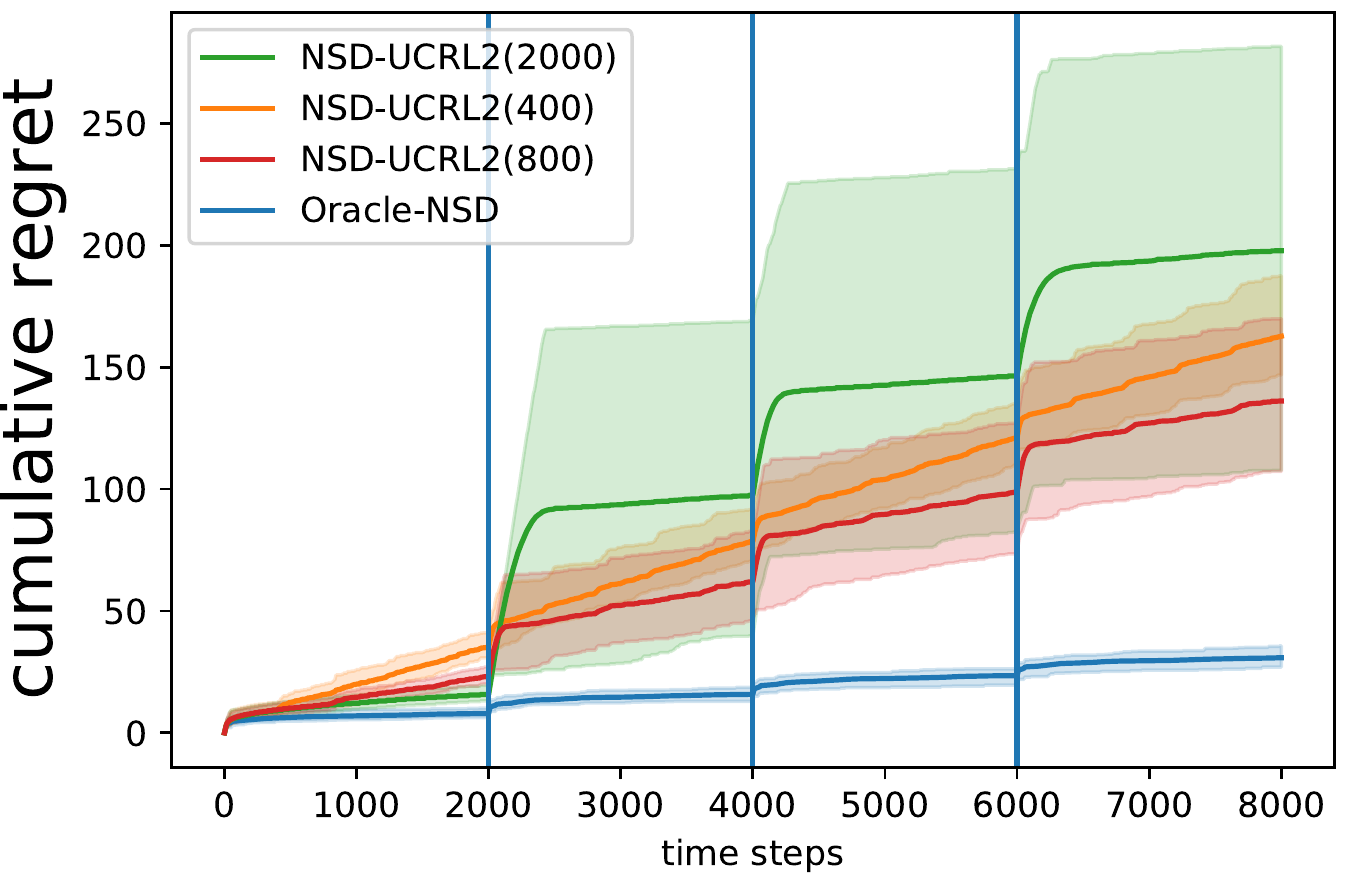}
     \vspace{-0.3cm}
    \caption{Impact of the choice of window parameter $W$ on the regret of $\NSDUCRL$. }
    \label{fig:window_exp}
     \vspace{-0.4cm}
\end{figure}

\paragraph{Choice of $W$.}
Our theoretical analysis (Theorem~\ref{th:main_regret}) suggests that for a problem with $T=8000$ and $\Gamma_T=3$, the optimal choice of the window parameter $W$ is of order $T^{2/3}\Gamma_T^{1/3}(KS)^{1/3}=1243$.
In Figure~\ref{fig:window_exp}, we compare the regret of $\NSDUCRL$ instantiated with $W\in \{400, 800, 2000\}$ to that of the oracle that knows the change points in a changing environment with no delays ($D=0$). We can make several observations on those results. First, when the window is too small, the algorithm never reaches the exploitation phase and the regret is linear.
This corresponds to the regime where the term in $T\sqrt{\log(WT)/W}$ (or $T\log T/(W\Delta_{\min})$ in Theorem~\ref{th:problem-dependent}) is dominant in the regret bound. 
Then, the larger the window gets, the better is the exploitation and the smaller the regret. However, for $W>1000$, the variance of the regret gets much larger as the overlap between the phases creates bias in the estimators. This roughly corresponds to the second regime where the term $W\Gamma_T$ becomes dominant. In the next experiments, we will set $W=800$, which is close to the value suggested by theory and that works best in this calibration experiment.

\paragraph{Comparison with all the baselines.}
 \begin{figure*}[!th]
 \vspace{-0.2cm}
     \centering
     \includegraphics[width=0.3\textwidth]{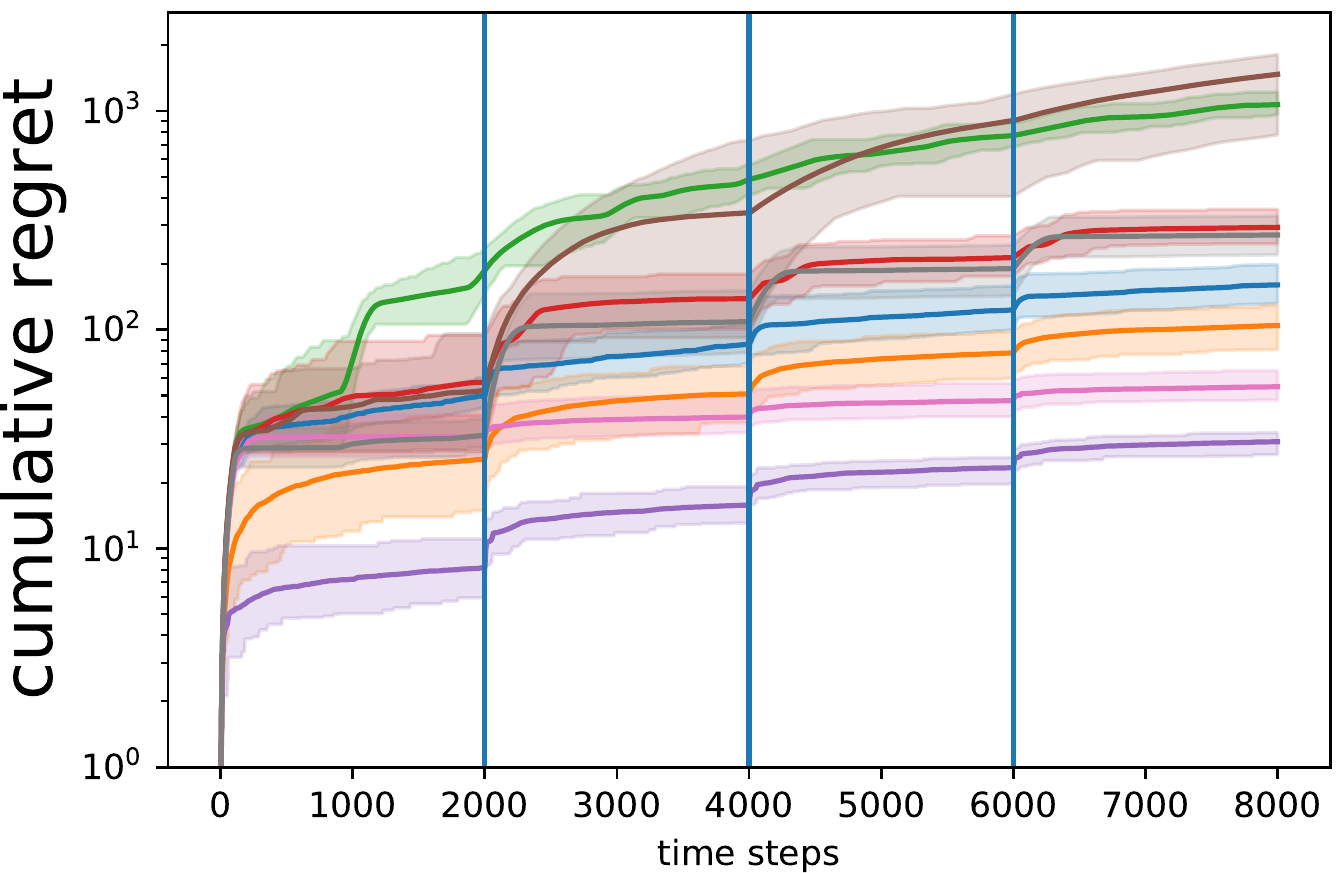}
     \includegraphics[width=0.28\textwidth]{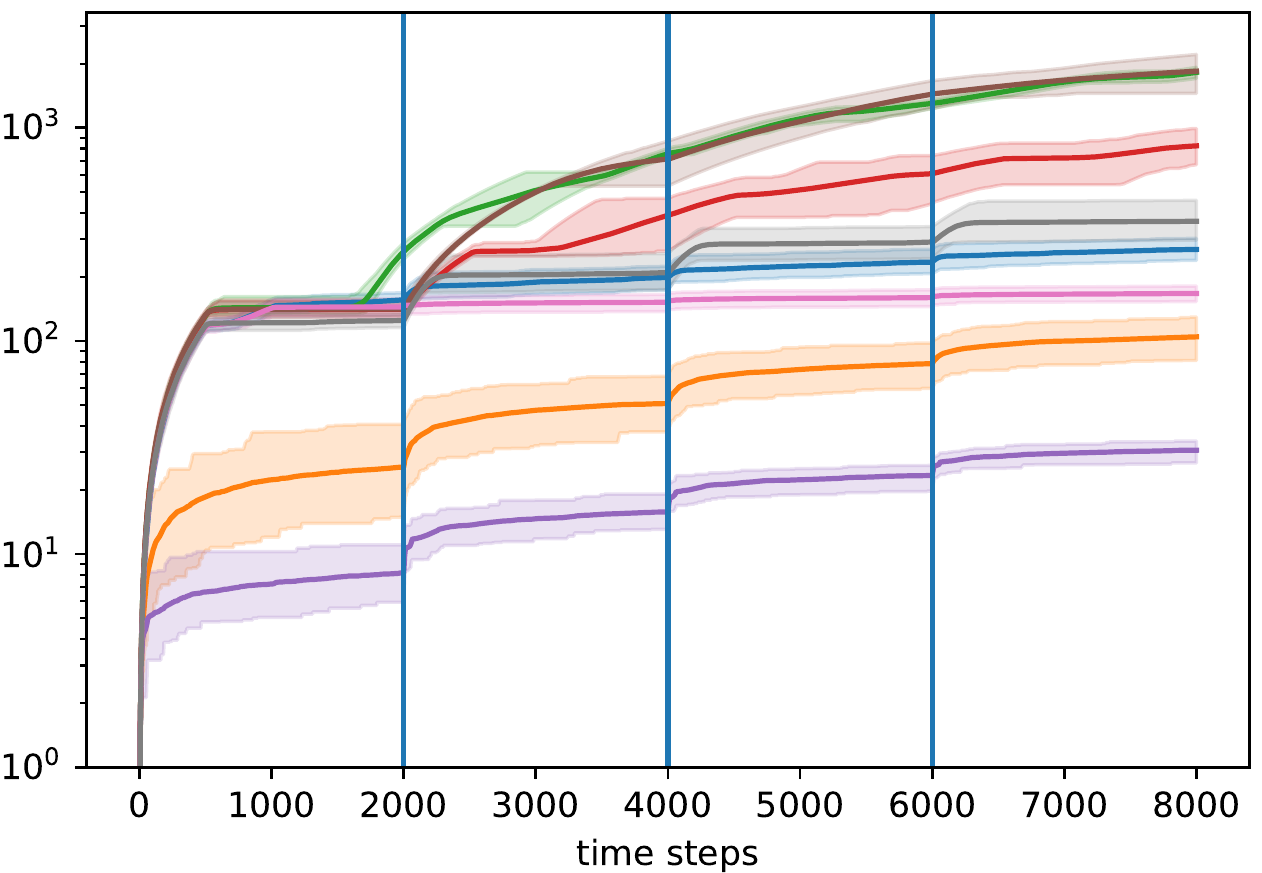}
     \includegraphics[width=0.37\textwidth]{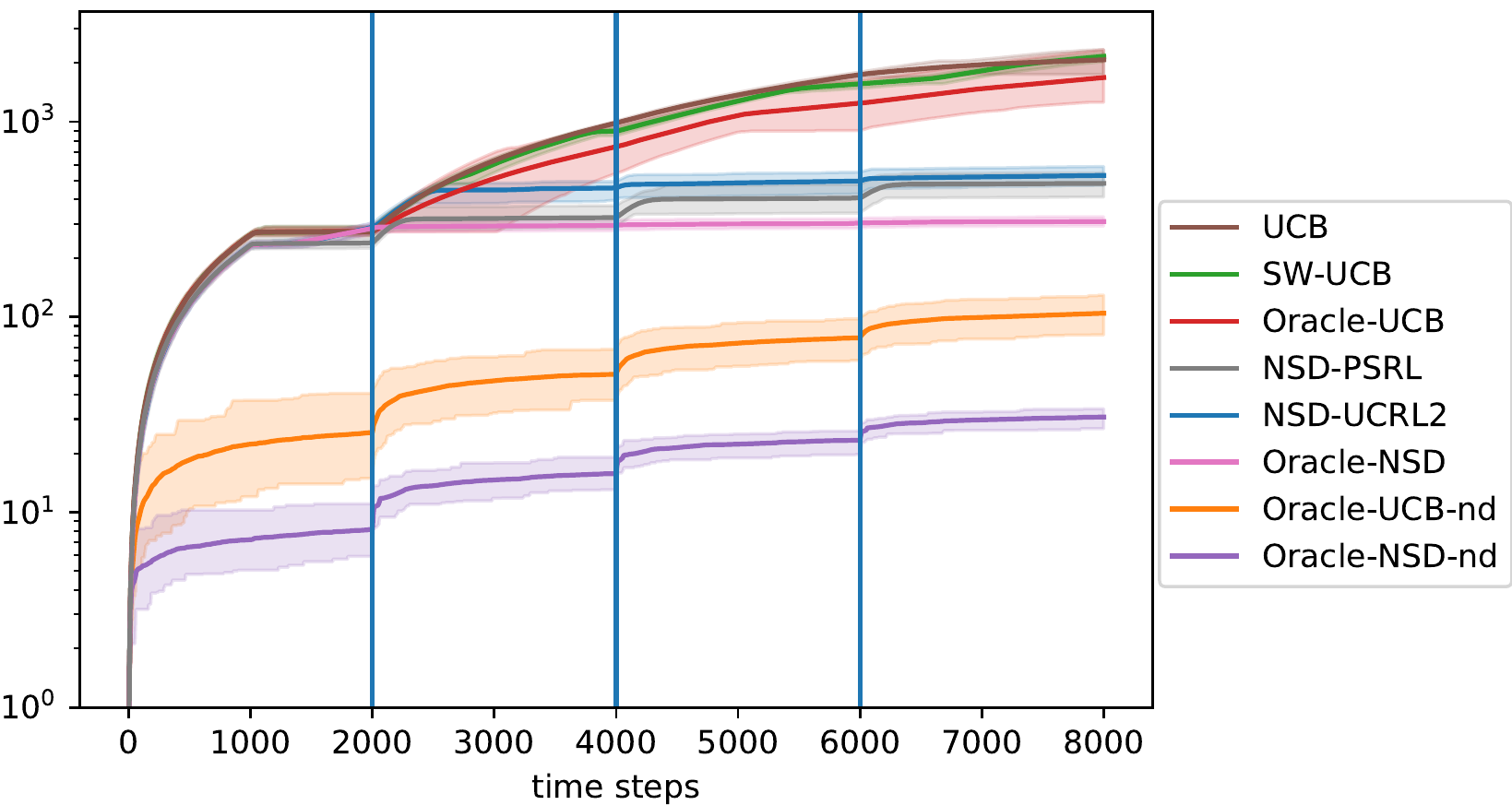}
       \vspace{-0.4cm}
     \caption{Cumulative regret of all policies in vertical log scale. $\Gamma_T=3$ and $D\in \{100, 500, 1000\}$ from left to right. When a policy uses a window, $W=800$. Oracles know the change points, non-delayed oracles (nd) receive rewards without delays.  }
     \label{fig:all_policies}
 \end{figure*}
 In this experiment, we set
 $T=8000$, $\Gamma_T = 3$, and consider delays $D\in\{100;500; 1000\}$. 
 The largest value corresponds to the extreme case of $D\approx T/\Gamma_T$. 
 For this experiment, we fix the window  parameter to $W=800$, for both $\SWUCB$ and $\NSDUCRL$.  

 Figure~\ref{fig:all_policies} shows the regret of all the policies in the benchmark, including strong oracles. Our $\NSDUCRL$ and $\NSDPSRL$ outperform all other strategies, except for the oracle ones. Non-delayed oracles have a significantly reduced regret but our policies have a similar behavior with a reasonable gap when delays are small. Signal-agnostic policies ($\UCB$ and $\SWUCB$) have a linear regret (see Appendix~\ref{ap:figures} for linear scale plots) as well as the strong baseline $\OrUCB$ when delays are large.

\paragraph{Ablation study: behavior under a misspecified model. }

Our factored reward model is likely not correct in practice. However, if the model is approximately accurate, we expect that our algorithm performs well. To examine such situations, we construct a mixture reward model where the factored model is correct only in some fraction of the time, while the rewards and the intermediate observations may behave differently at other times. 
For the latter, the rewards of the actions are controlled by an additional vector of parameters $\mu=(\mu_1, \ldots, \mu_K)$, and are assumed to be independent of the intermediate observations. Then, in each round $t$, given action $A_t\in [K]$, the environment either uses our factored model (with probability $1-\alpha$ for some mixing parameter $\alpha \in [0,1]$), or returns a random intermediate observation while generating a reward with expectation $\mu_{A_t}$. 
Formally,
\begin{align*}
    \text{with probability } \alpha, & \quad S_t \sim \cU ([S]) \text{ and } R_t \sim \cB(\mu_{A_t}); \\
    \text{with probability } 1- \alpha, & \quad S_t \sim p_t(\cdot|a) \text{ and } R_t \sim \cB(\theta_{S_t});
\end{align*}
where $\cU([S])$ denotes a uniform distirbution over $[S]$ and $\cB(\beta)$ denotes a Bernoulli distribution with parameter $\beta$.
This means that the expected payoff of arm $a$ at round $t$ is $\rho_\alpha (a) = \alpha \mu_a + (1-\alpha) p_t(\cdot |a)^\top \theta$. 

First we test how $\NSDUCRL$ performs in this environment in the stationary non-delayed setting, then we analyze its performance in the delayed non-stationary case (note that as long as the algorithm learns reasonably well in the stationary non-delayed case, we can expect to see a similar performance boost in the delayed non-stationary scenario due to the intermediate observations as in the case when the factored model is correct).

\textbf{In the stationary setting}, we identify two types of regime. In a favorable situation,  the best arm is not modified by the mixing, i.e. $\argmax_a \rho_\alpha(a)=\argmax_a \rho(a)$. In that case, we conjecture that $\NSDUCRL$ should be able to have a sublinear regret as it tends to learn the right action. On the contrary, in a bad mixing case, the best arm is not the same and we expect $\NSDUCRL$ to fail. In any case, $\UCB$ ignores the signals and should have a logarithmic regret. 

To test these properties, we set up a simple, stationary problem with only actions 1 and 2 from Table~\ref{tab:exp_setting}, such that action 1 is the best one. A favorable case is when $\mu=(0.9, 0.1)$ such that for any $\alpha$, the optimal action in the mixed model is action 1. A bad case is for instance when $\mu=(0.1,0.9)$, in which case the best arm is preserved only for small values of $\alpha$. We show some of our results below, more  can be found in Appendix~\ref{ap:figures}.

\begin{figure}
    \centering
    \includegraphics[width=0.5\columnwidth]{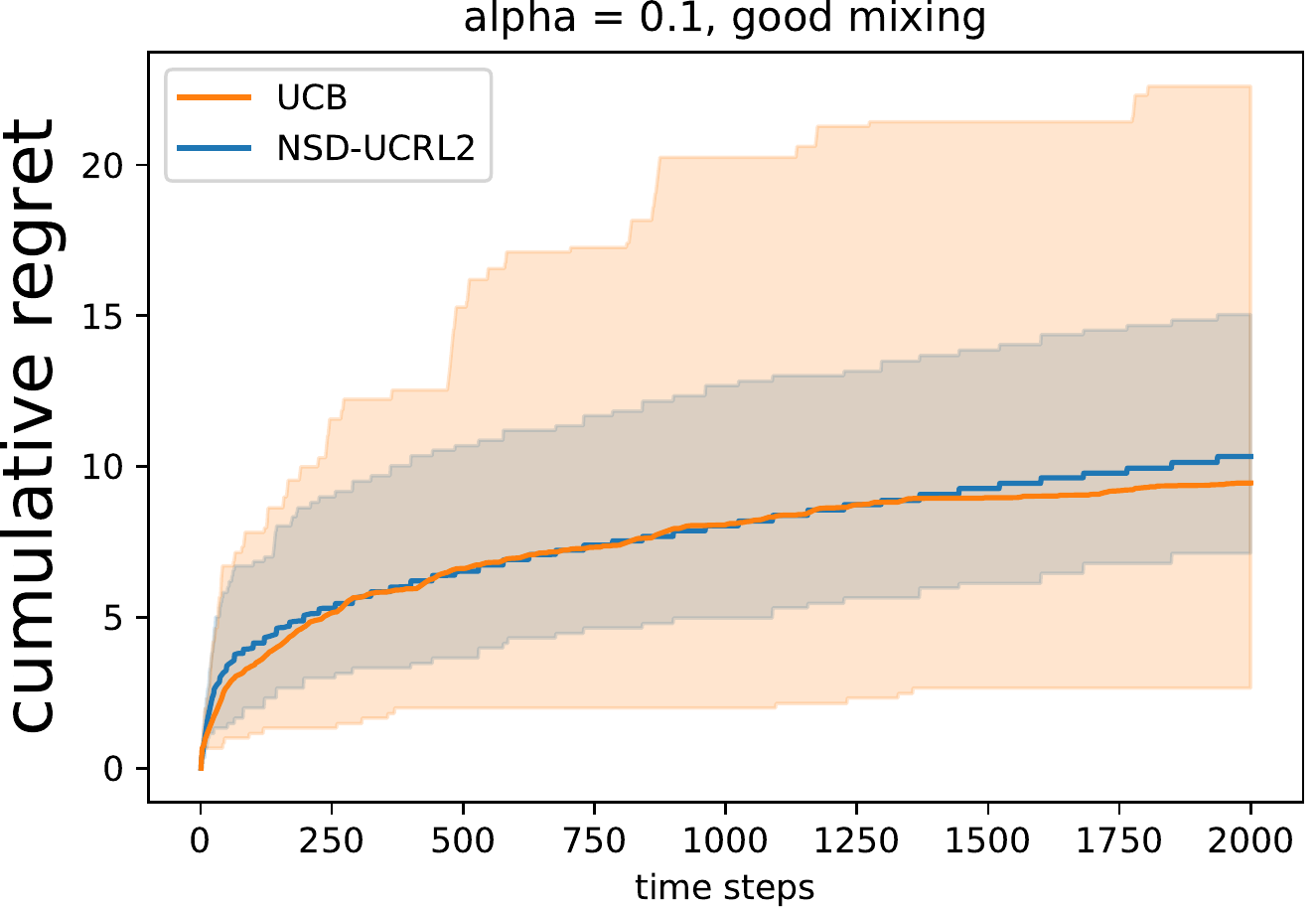}
    \includegraphics[width=0.47\columnwidth]{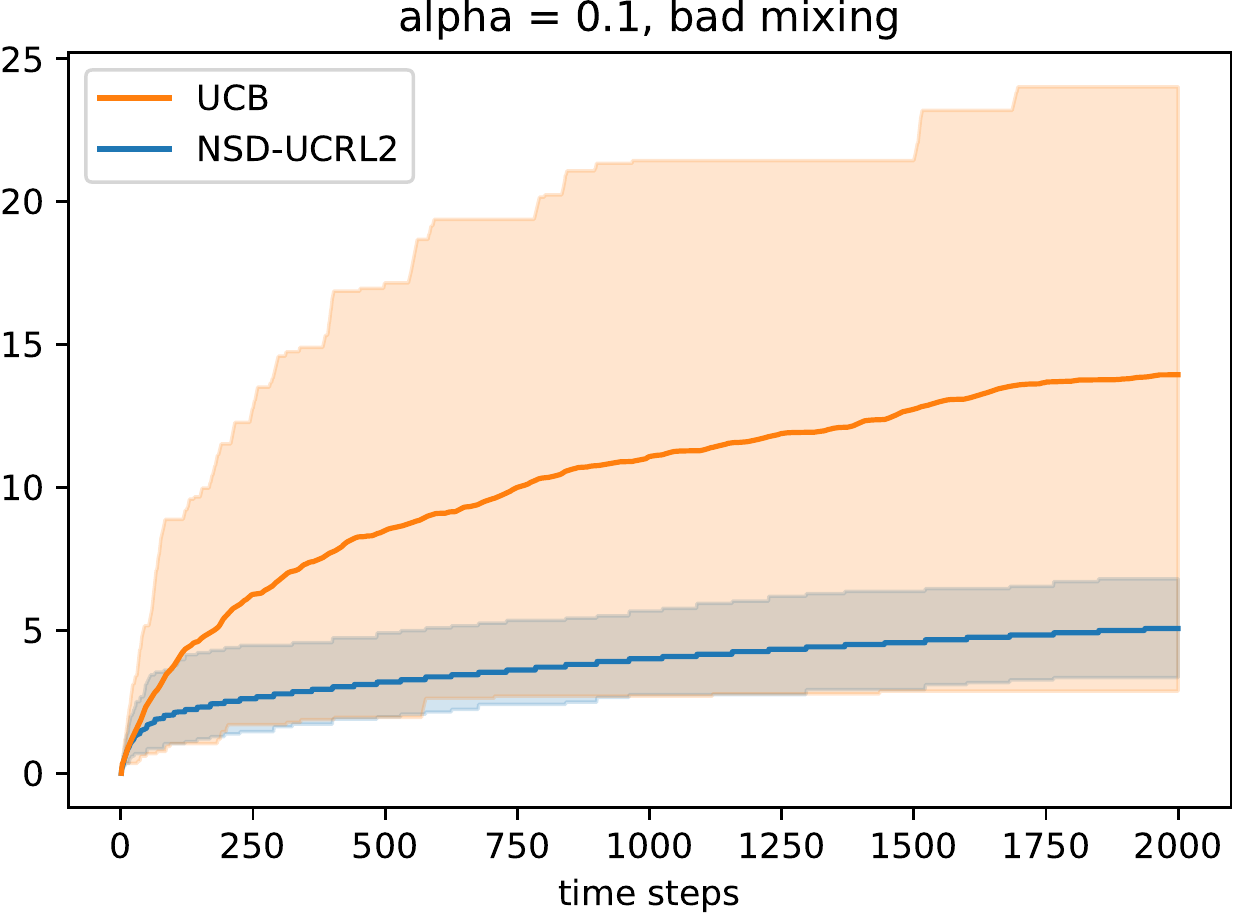}
    \vspace{-0.3cm}
    \caption{Misspecified model with low mixing level $\alpha=0.1$ in the stationary setting: In both favorable and bad cases, $\NSDUCRL$ is on par with  $\UCB$. The bad mixing case tends to increase the variance of the rewards and even increases $\UCB$'s regret. }
        \vspace{-0.5cm}
    \label{fig:low_mixing}
\end{figure}

\begin{figure}
    \centering
    \includegraphics[width=0.5\columnwidth]{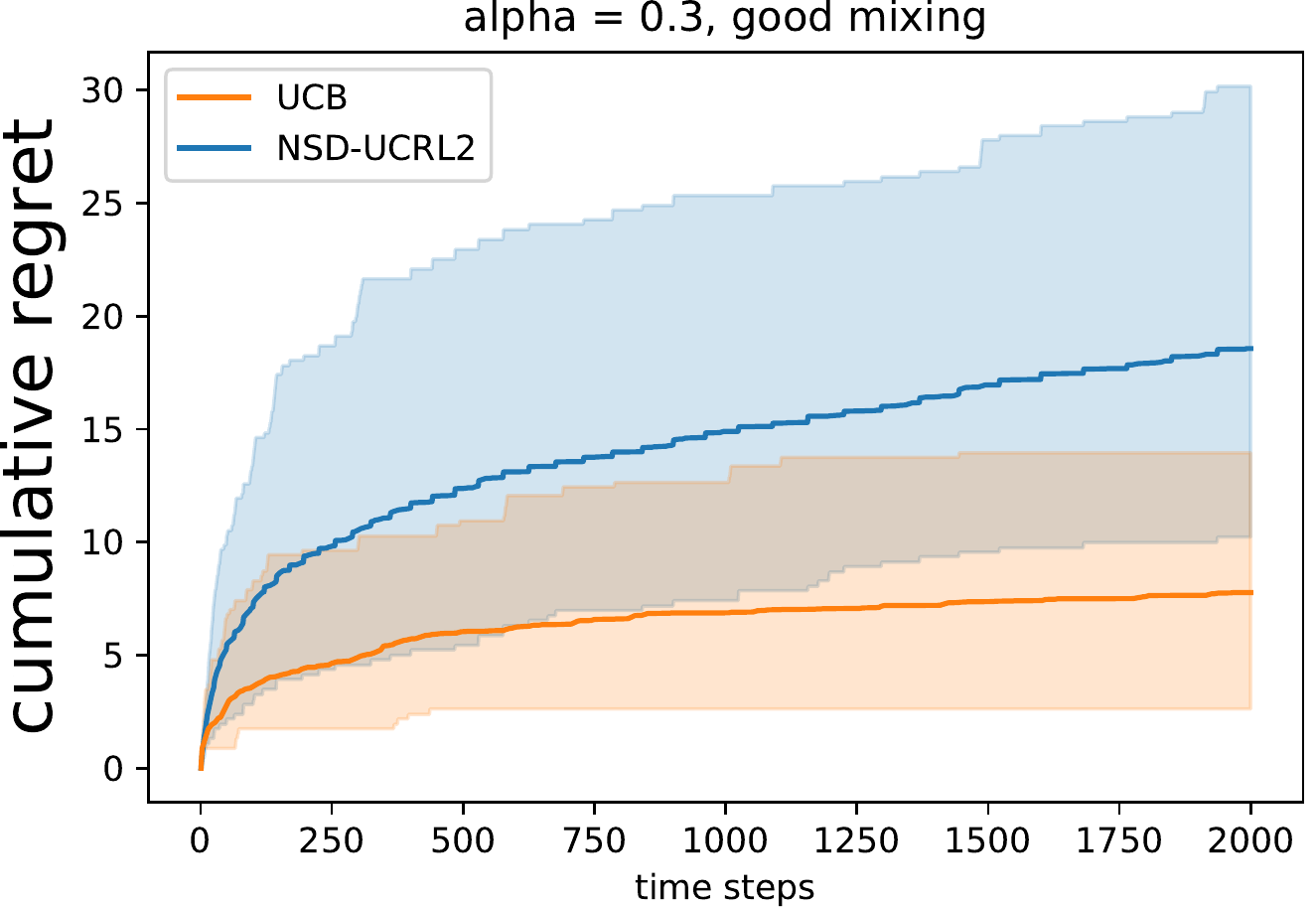}
    \includegraphics[width=0.47\columnwidth]{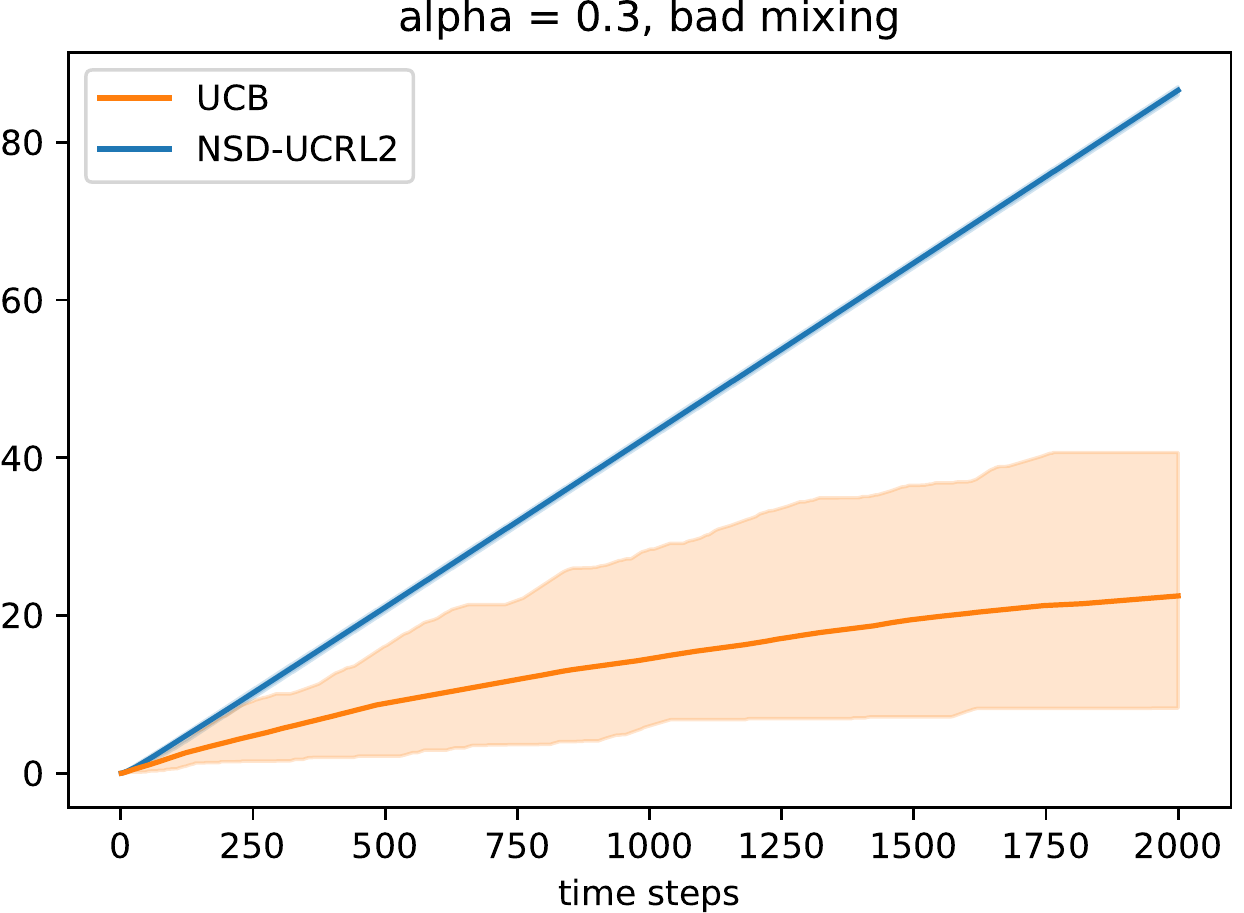}
    \vspace{-0.3cm}
    \caption{Misspecified model with non-negligible mixing level $\alpha=0.3$ in the stationary setting: $\NSDUCRL$ is on par with $\UCB$ in the favorable case, and diverges in the bad case. }
        \vspace{-0.5cm}
    \label{fig:decent_mixing}
\end{figure}

In Figure~\ref{fig:low_mixing}, we show the results in both scenarios in the case of a low mixing level, that is, when the model is close to right. $\NSDUCRL$ is on par or better than $\UCB$. $\UCB$ even suffers from the higher variance of the distribution of the rewards, especially in the bad mixing case. 
On the contrary, as expected, when the mixing becomes non-negligible ($\alpha=0.3$), $\UCB$ learns in both the favorable and bad cases. $\NSDUCRL$  also learns, though with higher regret in case of good mixing, but it fails in the bad case. 

\begin{figure*}
    \centering
    \includegraphics[width=0.32\linewidth]{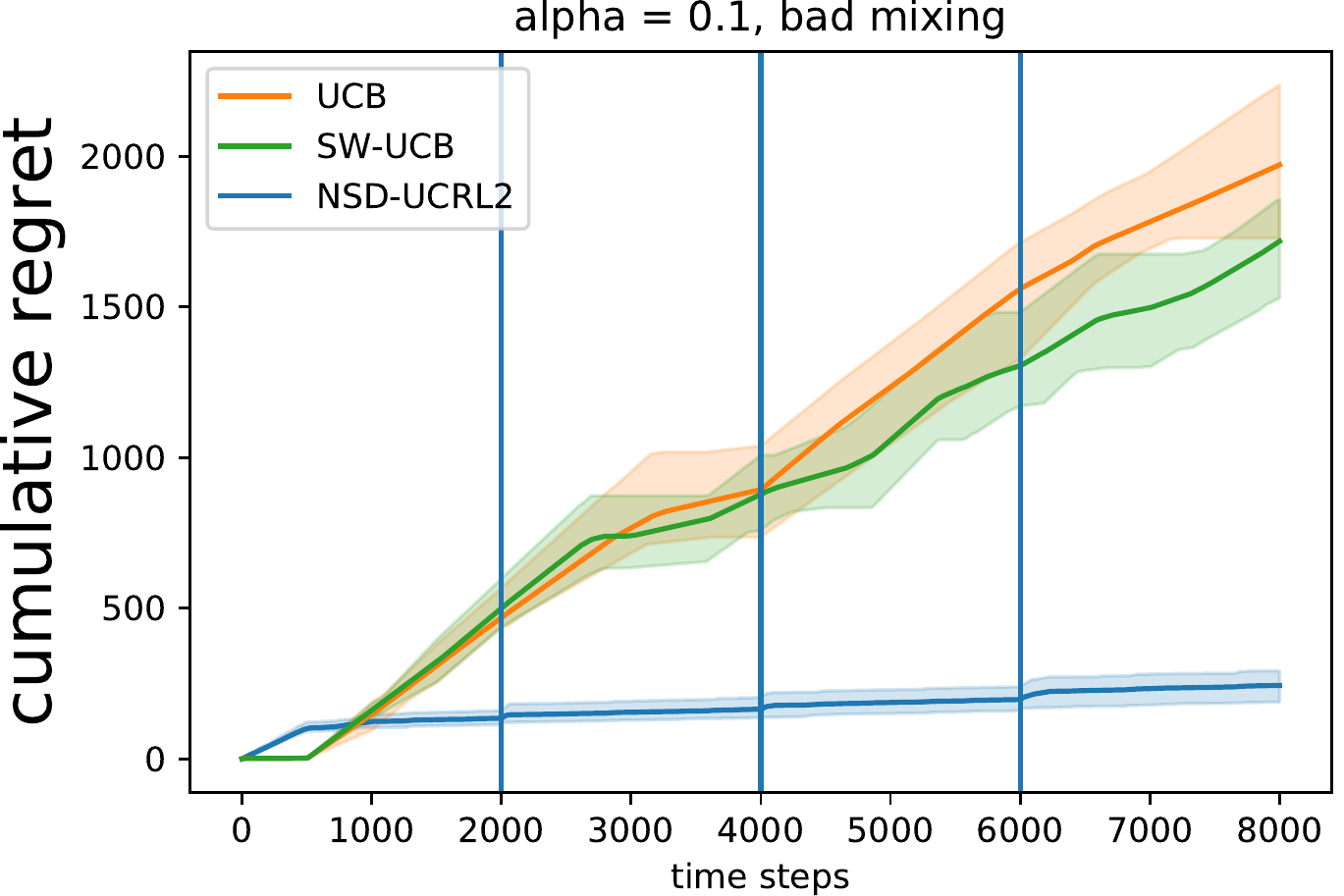}
    \includegraphics[width=0.32\linewidth]{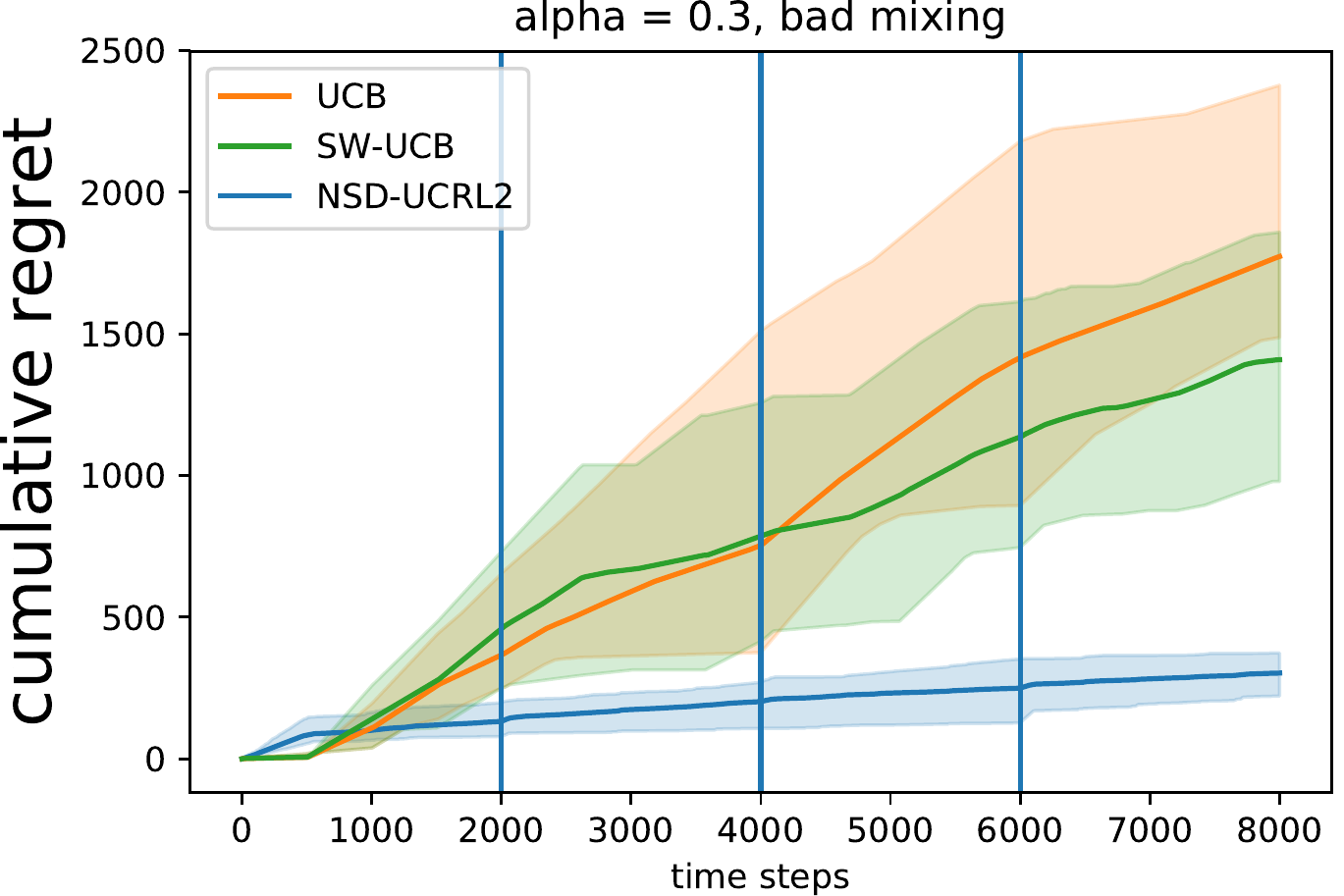}
    \includegraphics[width=0.32\linewidth]{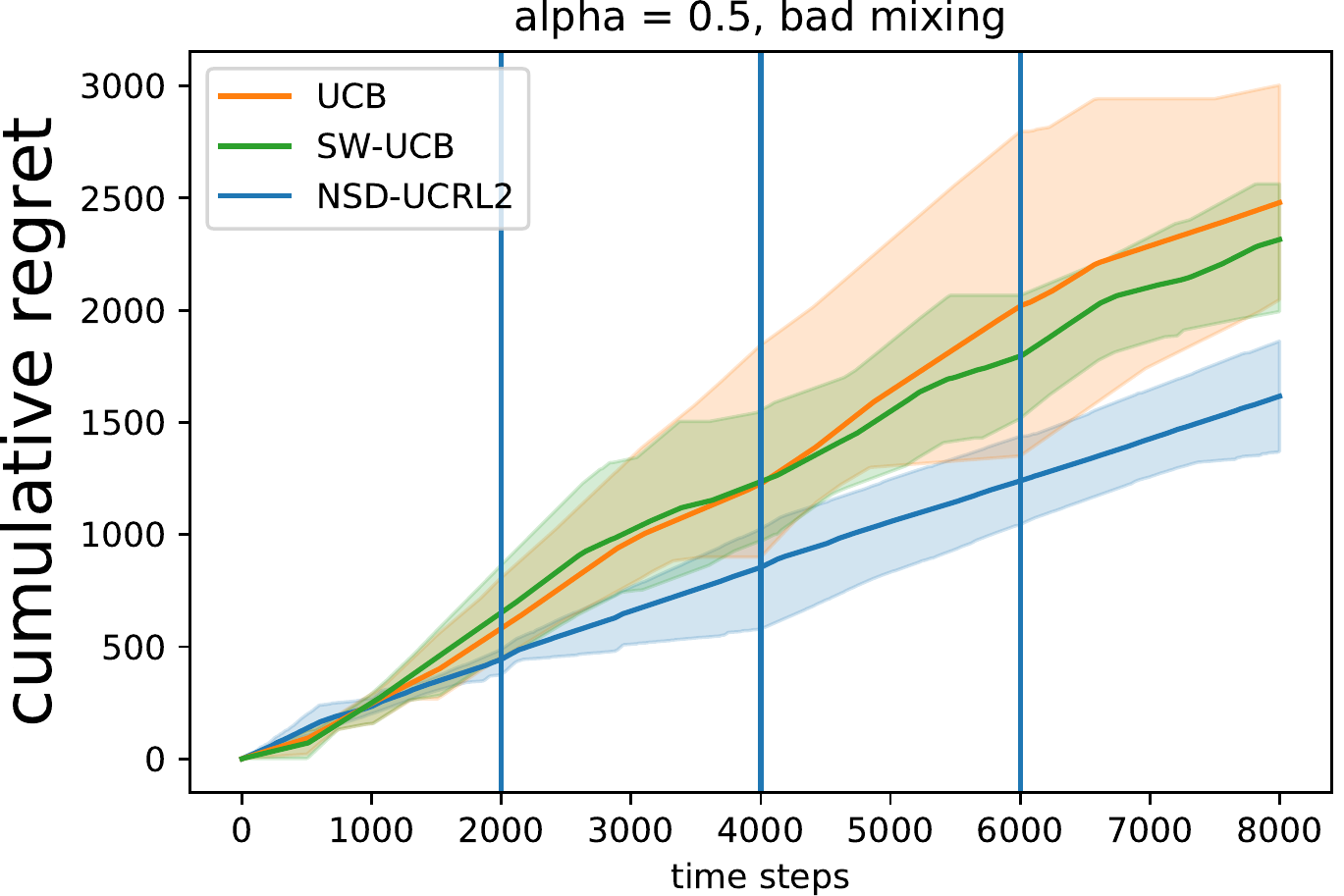}
    \caption{Regret of $\NSDUCRL$, $\SWUCB$ and $\UCB$ in the same setting as Figure~\ref{fig:all_policies} when the model is misspecified. Delays are fixed to $D=500$. From left to right, $\alpha=(0.1, 0.3, 0.5)$ and $\mu=(0.1,0.1,0.1,0.9)$ is chosen so that the best arm changes in the mixed model. Despite the mixing, even for non-negligible mixing levels, $\NSDUCRL$ learns decently fast. }
    \label{fig:nsd_mixed}
\end{figure*}

\textbf{In the non-stationary and delayed setting} considered before, the transition probabilities change along the horizon, together with the vector $\mu_t$ (now depending on time using the same shifting scheme as for the transition probabilities), modifying the means of the arms in a consistent manner. We compare the performance of $\UCB$, $\SWUCB$ and $\NSDUCRL$ when the model is misspecified. Unlike what was done before, we do not expect $\NSDUCRL$ to gracefully adapt to changes. Nonetheless, results in Figure~\ref{fig:nsd_mixed} show that $\NSDUCRL$ does learn when $\alpha$ is small, and significantly outperforms both $\UCB$ and $\SWUCB$. For $\alpha \ge 0.5$, none of the considered algorithms is able to solve the problem.

\section{Other Related Work}
\label{sec:related}

As mentioned in the introduction, our model combines the challenges of non-stationary stochastic bandits \cite{garivier2011upper, auer2019achieving} and those of delayed feedback in online learning \cite{joulani2013online}, and online learning in episodic MDPs \cite{DannLB17}.
The general tree-structure of the problem shown in Figure~\ref{fig:mdp} is also reminiscent of combinatorial (semi-)bandits \cite{cesa2012combinatorial,combes2015combinatorial,kveton2015tight,talebi2017stochastic}. However, the difference is that while in the aforementioned combinatorial bandit problems one can directly select which sub-actions to choose, in our model we cannot directly choose to observe the reward corresponding to a selected signal. This limits the adaptation of the techniques used in these algorithms, as the corresponding observation counts can only be controlled indirectly. %

\section{Conclusions}
\label{sec:conlusion}

In this paper, we defined a new stochastic bandit model, NSD bandits with intermediate observations. It addresses real-world modelling issues when dealing with non-stationary environments and delayed feedback, and mitigates the joint effect of these by utilizing some non-delayed proxy feedback (similarly to the work of \citealt{mann2018learning}). From the theoretical point of view, it provides a mid-way model between the simple stochastic multi-armed bandit and the tabular MDPs. %

One first possible future extension would be to consider the contextual version of this problem, where the transition probabilities of each action depends on a current context vector. Another potential extension would be to add several layers of signals that would be sequentially sent to the learner before the final reward is revealed. This problem would then resemble combinatorial bandits and episodic MDPs with varying delays. 

While we presented sub-linear regret guarantees, understanding the rate of the minimax regret and modifying the algorithm to achieve it is of obvious interest. Future work should also look into problem-dependent guarantees. For instance, it would be interesting to exploit ideas from \cite{graves1997asymptotically} to obtain a gap-dependent asymptotic lower bound, and seek optimal strategies.

\ifdefined\isaccepted
\section*{Acknowledgements}
The authors are indebted to Tor Lattimore for several inspiring discussions. 
\fi

\bibliographystyle{icml2020}
\bibliography{references}

\newpage
\onecolumn

\appendix

\input{appendix.tex}

\end{document}

%% file: header.tex
\usepackage{amsmath,amssymb,amsthm,dsfont,graphicx,bm}

\usepackage[dvipsnames]{xcolor}

\usepackage{xspace}
\usepackage[disable]{todonotes}

\usepackage{algorithm}
\usepackage{listings} %

\definecolor{dkblue}{cmyk}{1,.54,.04,.19}
\definecolor{dkgray}{rgb}{0.3,0.3,0.3}
\usepackage{hyperref}
\hypersetup{
      unicode=false,          
      pdftoolbar=true,        
      pdfmenubar=true,        
      pdffitwindow=false,     
      pdfstartview={FitH},    
      pdftitle={Non-stationary Delayed Bandits with Intermediate Observations},    
      pdfauthor={Vernade, Gy\"orgy, Mann},     
      pdfcreator={pdflatex},   
      pdfproducer={Producer}, 
      pdfnewwindow=true,      
      colorlinks=true,        
      linkcolor=dkgray,       
      citecolor=dkgray,       
      filecolor=dkblue,       
      urlcolor=dkblue,        
}

\usepackage{dsfont}

\usepackage[capitalize]{cleveref}

\newtheorem{theorem}{Theorem}

\newtheorem{proposition}{Proposition}

\newtheorem{remark}{Remark}

\DeclareMathOperator*{\argmax}{arg\,max\,}

\mathchardef\mhyphen="2D

\newcommand{\cB}{\mathcal{B}}
\newcommand{\cE}{\mathcal{E}}

\newcommand{\cU}{\mathcal{U}}

\newcommand{\cT}{\mathcal{T}}

\newcommand{\R}{\mathds{R}}

\newcommand{\F}{\mathcal{F}}

\renewcommand{\P}{\mathds{P}}
\newcommand{\ind}{\mathds{1}}

\newcommand{\E}{\mathds{E}}
\newcommand{\hp}{\hat{p}}
\newcommand{\tp}{\tilde{p}}

\newcommand{\UCB}{{\tt UCB}}

\newcommand{\SWUCB}{{\tt SW \mhyphen  UCB}}

\newcommand{\WUCRL}{{\tt SW \mhyphen UCRL2}}
\newcommand{\NSDUCRL}{{\tt NSD \mhyphen UCRL2}}
\newcommand{\NSDPSRL}{{\tt NSD \mhyphen PSRL}}
\newcommand{\PSRL}{{\tt PSRL}}
\newcommand{\UCRL}{{\tt UCRL2}}
\newcommand{\OrUCB}{{\tt Oracle \mhyphen UCB}}
\newcommand{\OrNSD}{{\tt Oracle \mhyphen NSD}}
\newcommand{\OrUCBnd}{{\tt Oracle \mhyphen UCB \mhyphen nd}}
\newcommand{\OrNSDnd}{{\tt Oracle \mhyphen NSD \mhyphen nd}}
\newcommand{\UBEV}{{\tt UBEV}}

\newcommand{\Reg}{\mathcal{R}}
\newcommand{\eps}{\varepsilon}
\newcommand{\ebad}{\cB^\eps}
\newcommand{\pmse}{p^\eps_{\min}(s)}

\usepackage{tikz}
\usetikzlibrary{calc,shapes,arrows}

%% file: appendix.tex
\section{Extended Value Iteration}
\label{ap:evi}
The pseudo-code in Algorithm~\ref{alg:EVI} describes how to solve 
\[
\rho^+_t(a)
= \max_{q \in \Delta_S} \left\{q^\top U_t : \|\hat{p}_t(a) -q \|_1 \leq \sqrt{\frac{C_{W,T,\delta}}{N^W_t(a)}}\,\right\}
\]
for every arm $a$ in each iteration. Our pseudo-code is essentially identical to Figure~2 of \citet{jaksch2010near}, and we only report it here for completeness.

\begin{algorithm}%
\caption{Computing $\rho^+_t(a)$ \label{alg:EVI}}
\begin{algorithmic}
\STATE{\textbf{Input}: transition probability estimate $\hat{p}=\hat{p}_{t}(a)$, tolerance parameter $\beta=\sqrt{\frac{C_{W,T,\delta}}{N^W_t(a)}}$, reward upper bounds $U=U_t$.}
\STATE{ \textbf{Initialization}: By ordering the entries of $U$, compute a permutation $s_1,\ldots,s_S$ of $[S]$ such that $U(s_1) \ge \ldots \ge U(s_S)$.}
\STATE{Set $q(s_1)=\min\{1,\hat{p}(s_1)+\beta/2\}$.\qquad\quad} \COMMENT{Saturate constraint for the signal with the largest value.} 
\STATE{Set $q(s_j)=\hat{p}(s_j)$ for $2 \le j \le S$.}
\STATE{j=S.}
\WHILE{$\sum_{i \in [S]} q(s_i)>1$}
\STATE{$q(s_j) = \max\{0, 1 - \sum_{i \neq j} q(s_i)\}$.\qquad}
\COMMENT{Progressively saturate constraints for the other signals.}
\STATE{$j=j-1$.}
\ENDWHILE
\STATE{\textbf{return }$\rho^+=q^\top U$.}
\end{algorithmic}
\end{algorithm}

\section{Proof of Theorem~\ref{th:stationary_regret}}
\label{ap:stationary_regret}

\textbf{Step 1: Low probability failure events.}
We first define three types of failure events:
\begin{align}
    \label{eq:E_0}
    \cE_0 & = \ind\Big\{ \exists\, t\in [T],\, \exists a\in [K],\, \, \rho^+_t(a) <\rho_t(a) \Big\}. \\
    \label{eq:E_1}
    \cE_1 &= \ind\left\{\exists t \in [T], \exists a\in[K], \|\hat{p}_t(a)-p_a\| \geq \sqrt{\frac{C_{W,T,\delta}}{N^W_T(a)}} \right\} \\
    \cE_2 & = %
    \ind \left\{ \exists t\in [T], s\in [S], \, |\hat{\theta}(s)-\theta_s| > \sqrt{\frac{C_{T,\delta}}{N^D_t(s)}} \,\right\}
    \nonumber
\end{align}
In Appendix~\ref{ap:failure_event} we prove that
\begin{equation}
    \label{eq:events}
\ind{P}(\cE_0 \cup \cE_1 \cup \cE_2)) \leq \ind{P}(\cE_1)+\ind{P}(\cE_2) \leq 2\delta.
\end{equation}
First it is shown that $\cE_1$ and $\cE_2$ together imply $\cE_0$, then the probability of the former two events are bounded using standard techniques.

\textbf{Step 2: Regret decomposition.} Now we bound the regret under the assumption that none of the events $\cE_0,\cE_1,\cE_2$ hold using the standard techniques for optimistic algorithms:
\begin{align*}
    & \Reg(T) = \sum_{t=1}^T \rho^*_t - \rho_{A_t,t} 
        \leq \sum_{t=1}^T \rho^+(A_t) - \rho_{A_t,t} \\
        & = \sum_{t=1}^T \tilde{p}_t(A_t)^\top U_t - p_t(A_t)^\top \theta \\
        & = \underbrace{\sum_{t=1}^T (\tilde{p}_t(A_t) - p_t(A_t))^\top U_t}_{\text{A}} + \underbrace{\sum_{t=1}^T p_t(A_t)^\top (U_t -\theta)}_{\text{B}} 
\end{align*}
It remains to bound each term individually.
Since $\cE_1$ does not occur, we have
\[
A \le \sum_{t=1}^T \| \tp_t(A_t) - p_t(A_t) \|_1 \|U_t\|_\infty
\le \sum_{t=1}^T 2\sqrt{\frac{C_{W,T,\delta}}{N^W_t(A_t)}}~.
\]
To control the latter sum above, split the time horizon into intervals of length $W$; for any interval $I_l=[(l-1)W,lW-1]$ and any $t \in I_l$, let $N^l_t(a)$ denote the number of times action $a$ was chosen in $[(l-1)W,t-1]$, and set it to $1$ if it was not chosen. Then clearly $N_t^W(a) \ge N^l_t(a)$ for any $a$, and we have
\begin{equation}
    \label{eq:termA}
\sum_{t \in I_l} \frac{1}{N_t^W(A_t)} \le 
\sum_{t \in I_l} \frac{1}{N_t^l(A_t)} \le
2 \sqrt{K(W+1)}.
\end{equation}
Here the second inequality holds because the sum is the largest when each action is chosen $\lfloor W/K \rfloor \le W/K$ times in $I_l$ (up to rounding errors--note that since the first real count can be zero, the $1/\sqrt{1}$ terms are repeated twice for each action)%
and the inequality $\sum_{u=1}^v 1/\sqrt{u} \le 2(\sqrt{v+1}-1)$.
Considering that the number of intervals is $\lceil T/W \rceil \le T/W+1$, we obtain the bound
\begin{align*}
A \leq  O\!\left(T\sqrt{\frac{K C_{W,T,\delta}}{W}}\right)
= O\!\left( T\sqrt{\frac{S K \log(KWT/\delta)}{W}}\right).
\end{align*}

To bound term B, use that by the definition of $U_t$ and since $\cE_2$ does not occur, for any $s$ we have
\begin{equation}
    \label{eq:uts}
|U_{t,s} - \theta_s|
\le \sqrt{\frac{C_{T,\delta}}{N^D_t(s)}}~.
\end{equation}
However,  term B features the entanglement of action and signal-related terms.  For each signal $s$, the counts $N^D_t(s)$ are not directly increased by the actions of the agents but only randomly through the transition probabilities $p(A_t,s)$. Hence, we need to further decompose this sum. 
To this end, let $\F_t$ denote the $\sigma$-algebra generated by $A_t$ and all random variables available to the algorithm before round $t$, and let $e_s$ denote the $s$-th standard unit vector in $\R^S$. We can decompose the term B as 
\begin{equation}
\label{eq:B_decomposition}
    B = \sum_{t=1}^T e_{S_t}^\top (U_t-\theta) + (p_t(A_t) - e_{S_t})^\top (U_t-\theta)~.
\end{equation}
Then $e_{S_t}, A_t$ and $U_t$ are $\F_t$-measurable, $\E[e_{S_t}|\F_t]=p_t(A_t)$,
and $(e_{S_t}-p_t(A_t))^\top (U_t-\theta)$ is a martingale-difference sequence with respect to $(\F_t)$. Furthermore, for every $t \in [T]$ and $s \in [S]$, 
$|U_{t,s}-\theta_s|
\le |\hat{\theta}_s-\theta_s| + |U_{t,s} - \hat{\theta}_s| \le 1 +\sqrt{C_{T,\delta}}$, and so
\[
|(e_{S_t}-p_t(A_t))^\top (U_t-\theta)| \le 2+2\sqrt{C_{T,\delta}}~.
\]
Thus,
by the Hoeffding-Azuma inequality \cite{azuma1967weighted},
with probability at least $1-\delta$,
\begin{equation}
\label{eq:p-e}
\sum_{t=1}^T (p_t(A_t) - e_{S_t})^\top (U_t-\theta) \le (2+2\sqrt{C_{T,\delta}})\sqrt{2 T \log\left(\frac{1}{\delta}\right)}~,
\end{equation}
and we only need to bound
$\sum_{t=1}^T e_{S_t}^\top (U_t-\theta)$.
Looking at each coordinate separately, 
\eqref{eq:uts} implies that if $\cE_2$ does not occur, for each $s\in[S]$,
\begin{align}
\label{eq:ind1}
\sum_{t=1}^T \ind\{S_t=s\} (U_t(s) - \theta_s)
& \le \sum_{t=1}^T \ind\{S_t=s\} \sqrt{\frac{C_{T,\delta}}{N_t^D(s)}}~.
\end{align}
Defining $N_t(s):=N_{t+1}^{0}(s)=\sum_{u=1}^t \ind\{S_t=s\}$,
we have
\begin{align}
\sum_{t=1}^T \frac{\ind\{S_t=s\}}{\sqrt{N_t^D(s)}}
 = \sum_{t=1}^T \frac{\ind\{S_t=s\}}{\sqrt{N_t(s)}} %
+ \sum_{t=1}^T \ind\{S_t=s\}\left(\frac{1}{\sqrt{N_t^D(s)}} - 
\frac{1}{\sqrt{N_t(s)}}\right)~.
\label{eq:sqrtn}
\end{align}
Since $N_t(s)$ increases by $1$ every time $\ind\{S_t=s\}$, 
the first term on the right-hand side can be bounded as 
\begin{equation}
  \sum_{t=1}^T \frac{\ind\{S_t=s\}}{\sqrt{N_t(s)}} \le 
  \sum_{n=1}^{N_T(s)} \frac{1}{\sqrt{n}} \le 2\sqrt{N_T(s)}~.
  \label{eq:sqrtlemma}
\end{equation} 
To bound the second term, notice that $N_t^D(s) \ge N_t(s) - D - 1$, thus as long as the latter is positive,
\begin{align*}
\frac{1}{\sqrt{N_t^D(s)}} - 
\frac{1}{\sqrt{N_t(s)}}
 & \le
\frac{1}{\sqrt{N_t(s) - D -1}} - 
\frac{1}{\sqrt{N_t(s)}} %
\\
& = 
\frac{D+1}{\sqrt{N_t(s)(N_t(s)\!-\!D\!-\!1)}\left(\sqrt{N_t(s)}+\sqrt{N_t(s)\!-\!D\!-\!1}\right)} \\
& \le \frac{D+1}{2 (N_t(s)-D-1)^{3/2}}
\end{align*}
Therefore, since $N_t(s) \le D+1$ can only hold for $D+1$ times when the indicator in the second term of \eqref{eq:sqrtn} is non-zero (in which case we upper bound the summands by 1), the second term of \eqref{eq:sqrtn} can be bounded by
$5(D+1)/2$, where we used the fact that
$\sum_{t=1}^T t^{-3/2} \le 3$.

Combining with \eqref{eq:ind1}--\eqref{eq:sqrtlemma} (and since we can bound by 1 the first $D+1$ terms of \eqref{eq:ind1} when the indicator is non-zero), we have
\begin{align*}
\sum_{t=1}^T e_{S_t}^\top (U_t-\theta) &\leq \sum_{s=1}^S\sum_{t=1}^T \ind\{S_t=s\} (U_t(s) - \theta_s) \\
& \le 2\sqrt{C_{T,\delta}} \sum_{s=1}^S
\sqrt{N_T(s)} + 5 S (D+1)/2 \\
&\le 2 \sqrt{S T C_{T,\delta}} + 5S(D+1)/2 
\end{align*}
where in the last inequality we used Jensen's inequality and the fact that $\sum_{s=1}^S N_T(s) = T$.
This, together with \eqref{eq:p-e} implies that with probability at least $1-2\delta$ (i.e., under the events we assumed and the concentration inequality \eqref{eq:p-e} which holds with probability at least $1-\delta$)
\begin{equation}
     \label{eq:termB}
B \le O\left(\left(\sqrt{S}+1+\sqrt{\log(1/\delta)}\right)\sqrt{T\log\left(\frac{TS}{\delta}\right)} + S(D+1)
\right)~.
\end{equation}

\subsection{Bounding the probability of the failure event: proof of \eqref{eq:events}}
\label{ap:failure_event}
We prove that
\begin{equation}
\P(\cE_0 \cup \cE_1 \cup \cE_2)) \leq \P(\cE_1)+\P(\cE_2) \leq 2\delta.
\tag{\ref{eq:events}}
\end{equation}
First note that if $\cE_1$ and $\cE_2$ are both false, then (i) $p_t(a)$ belongs to the set of distributions over which the maximum is computed in \eqref{eq:rho}; and (ii) for any $q \in \Delta_S$, $q^\top U_t \ge q^\top \theta$ holds since $U_t(s) = \hat{\theta}_s + C_T \ge \theta_s$. Therefore, in this case, 
\begin{align*}
\rho^+_t = \tp_t(a)^\top U_t \ge p_t(a)^\top U_t \ge p_t(a)^\top \theta  = \rho_t,
\end{align*}
which means that $\cE_0$ is also false. Therefore, $\cE_1 \cup \cE_2$ imply $\cE_0$, proving the first inequality in \eqref{eq:events} (together with the union bound).

As usual in UCB-type proofs \cite{auer2002finite}, $\P(\cE_1)$ is bounded by taking the union bound over the events that the $L_1$ error of $\hp_t(a)$ in estimating $p_t(a)$ from $u \in [W]$ observations is larger than $\sqrt{C_{W,T,\delta}/u}$. Defining $\bar{p}_{t,u}(a)$ as the empirical estimate of $p_t(a)$ from taking the first $u$ transitions corresponding to action $a$ in the window $[t-W,t-1]$ (with imaginary samples added if needed; also note that $\hp_t(a)=\bar{p}_{t,N^W_t(a)}$), and using the well-known concentration bound of \cite{weissman2003inequalities}, given in Theorem ~\ref{th:weissman} in Appendix~\ref{ap:weissman}, we get
\begin{align}
    \P(\cE_1) &\le \sum_{t=1}^T \sum_{a=1}^K \sum_{u=1}^W 
    \P\left(\|\bar{p}_{t,u}(a) - p_t(a)\|_1 > \sqrt{C_{W,T,\delta}/u}\right) %
    \le  %
    \delta~.
    \label{eq:E1}
\end{align}

The probability of $\cE_2$ can be bounded similarly using the Hoeffding-Azuma inequality: defining $\bar{\theta}_{u}(s)$ as the averaging estimate based on the first $u$ observations for rewards corresponding to signal $s$ (again, $\hat{\theta}_t(s)=\bar{\theta}_{N^D_t(s)}$), the union bound implies
\[
\P(\cE_2) \le \sum_{u=1}^{T-D} \sum_{s=1}^S \P\left(|\bar{\theta}_{u}(s) -\theta_s| > \sqrt{\frac{C_T}{u}}\right)
\le \delta~.
\]

Together with \eqref{eq:E1}, this proves \eqref{eq:events}.

\subsection{Auxiliary technical results}
\label{ap:weissman}
We recall the following inequality from \cite{weissman2003inequalities} which provides a tight control of the deviations of an empirical probability vector to its true value.
\begin{theorem}
\label{th:weissman}
Let $Q$ be a probability distribution on the set $[S] = {1, . . . ,S}$. Let $X^n =
X_1, X_2, ..., X_n$ be independent identically distributed random variables distributed according to $Q$. The empirical estimate of $Q$ is defined for each $s\in[S]$ as 
\[
\hat{Q}_l(n) = \frac{1}{n} \sum_{t=1}^n \ind\{X_t=s \}. 
\]
Then, for any $\delta >0$, 
\[
P\left( \| \hat{Q}_l(n)-Q \|_1 \geq \sqrt{\frac{2S\log(2/\delta)}{n}}\, \right) \leq \delta.
\]
\end{theorem}

\section{Proof of Theorem~\ref{th:main_regret}}
\label{ap:main_proof}

There are $\Gamma_T+1$ stationary phases and we denote $1=\tau_0,\ldots \tau_{\Gamma_T},\tau_{\Gamma_T+1}=T$ the rounds when a new stationary phase starts. 
Let $\mathcal{T} (W) = \{1\leq t\leq T:\, \forall k\in [K], \, \forall t-W\leq s \leq t,\, p_s(k)= p_t(k) \}$ denote the round indices when all arms have stable transition probabilities for at least $W$ rounds. 

Redefining the error events $\cE_0$ and $\cE_1$ defined in \eqref{eq:E_0} and \eqref{eq:E_1} for $t\in \mathcal{T}(W)$ only instead of $t\in[T]$ (and denoting the resulting events by $\cE'_0$ and $\cE'_1$, resp.), we consider the failure event $\cE'_0 \cup \cE'_1 \cup \cE_2$.
The difference from the stationary problem is that estimation errors in the rounds where the changing transitions cannot be estimated reliably are not considered. Nonetheless, by the same argument as before, this event has a probability at most $2\delta$.

Assuming none of $\cE'_0, \cE'_1$ and $\cE_2$ hold and applying the regret decomposition of Step 2 in the proof of  Theorem~\ref{th:stationary_regret}, we obtain that with probability at least $1-2\delta$,
\begin{align*}
    R(T) 
      & \leq W\Gamma_T + \sum_{i=0}^{\Gamma_T} \sum_{t=\tau_i+W}^{\tau_{i+1}-1} \|\tp_t(a) - p_a\|_1 + \sum_{t=1}^T p_t(A_t)^\top (U_t(s)-\theta_s).
\end{align*}
    
Each inner sum in the first term can be bounded as in the proof of Theorem~\ref{th:stationary_regret}:
\[
\sum_{t=\tau_i+W}^{\tau_{i+1}-1} \!\!\|\tp_t(a)-p_a\|_1
\le O\!\!\left(\!\!(\tau_{i+1}-\tau_i)\sqrt{\frac{KS\log\left(\frac{KWT}{\delta}\right)}{W}}\,\right)\!\!.
\]
Summing up for all $i$, and bounding the second term by
$O((\sqrt{S}+1+\sqrt{\log(1/\delta)})\sqrt{T\log(TS/\delta)}+S(D+1))$ as in \eqref{eq:termB}
proves the theorem.

\section{Proof of Theorem~\ref{th:problem-dependent}}
\label{ap:prob_dep_bounds}

We use the notation from the proof of Theorem~\ref{th:main_regret}: Recall that $\cT=\cT(W)$ denotes the set of rounds from $[T]$ where $W$ rounds are excluded after every change point, and that $\cE_0, \cE_1, \cE_2$ are failure events for our estimates.

In the rest of the proof we assume that none of $\cE_0,\cE_1$ and $\cE_2$ hold, which happens with probability at least $1-\delta$.
Let $0 < \Delta < \eps$, and for every round $t$ let $\ebad_t$ denote the set of $\eps$-bad actions at time $t$. For every $s \in [S]$, define the set of rounds
\[
T_s = \{t \in [T]: A_t \in \ebad_t, p_t(s|A_t) >0, |U_t(s)-\theta_s| > \Delta\}~.
\]
Furthermore, let
\[
T_{\Delta} = \{ t\in [\cT]: A_t \in \ebad_t, |
U_t(s) - \theta_s| \le \Delta \text{ for all $s$ such that }p_t(s|A_t)>0\}~.
\]
Clearly, $\{t\in [\cT]: A_t \in \ebad_t\} \subset T_{\Delta} \cup \bigcup_s T_s$. Therefore, to bound the number of times an $\eps$-bad action is selected, it is enough to bound from above the size of the sets $T_s$ and $T_{\Delta}$.

First note that since $A_t$ is selected in round $t$ and $\theta_s \le U_t(s)$ under the complement of $\cE_2^c$,
\begin{align}
\label{eq:astar}
p_t(a^*_t)^\top \theta \le p_t(a^*_t)^\top U_t \le \tp_t(a^*_t)^\top U_t \le \tp_t(A_t)^\top U_t.
\end{align}
Furthermore, for any $t \in T_{\Delta}$,
\[
\tp_t(A_t)^\top U_t \le \tp_t(A_t)^\top \theta + \Delta
\le p_t(A_t)^\top \theta + \Delta + (\tp_t(A_t) - p_t(A_t))^\top \theta \le p_t(A_t)^\top \theta + \Delta + \sqrt{C_{W,T,\delta}/N^W_t(A_t)},
\]
where in the last step we used that $\theta_s \in [0,1]$ and the definition of $\tp_t$.
Combining the above with \eqref{eq:astar}, we get
\[
p_t(a^*_t)^\top \theta - p_t(A_t)^\top \theta - \Delta \le \sqrt{C_{W,T,\delta}/N^W_t(A_t)}.
\]
Since $A_t$ is an $\eps$-bad action at time $t$ and $\Delta<\eps$,  
\[
N^W_t(A_t) \le \frac{C_{W,T,\delta}}{(\eps- \Delta)^2}.
\]
Covering $\cT$ with at most $\lceil T/W \rceil$ windows of size at most $W$ each, it follows that each of these windows can intersect $T^a_\Delta=\{t \in T_\Delta: A_t=a\}$ in at most $\frac{C_{W,T,\delta}}{(\eps- \Delta)^2}$ points for each $a \in [K]$. 
Therefore,
\begin{align}
    \label{eq:TDelta}
    |T_\Delta| \le \left\lceil \frac{T}{W} \right\rceil \frac{K C_{W,T,\delta}}{(\eps- \Delta)^2}
\end{align}

Next we bound the size of $T_s$ for every $s\in[S]$. Fix $s$. Since under our good events, 
$0 \le U_t(s) - \theta_s \le 2\sqrt{\frac{C_{t,\delta}}{N^D_t(s)}}$,
$|U_t(s) - \theta_s| > \Delta$ implies 
\begin{equation}
\label{eq:NDt-bound}
    N^D_t(s)< 4 C_{t,\delta}/ \Delta^2.
\end{equation}

Let $T_{s,D}=\{t \in T_s: t \le \max T_s -D\}$. Note that any action taken in $T_{s,D}$ receives a reward feedback by the end of $T_s$, and let $N_s$ denote the number of observations from $s$ received for rounds belonging to $T_{s,D}$. 

Since in every round $t\in T_{s,D}$, with probability at least $\pmse$ we get a sample from signal $s$, by the Chernoff bound (and the union bound over the size of $T_{s,D}$ and all $s \in [S]$) we get that if $|T_{s,D}|>\frac{8}{\pmse}\log\frac{TS}{\delta \pmse}$ then, with probability at least $1-\delta$,
\[
N_s > \pmse |T_{s,D}|/2.
\]
Thus, with high probability, \eqref{eq:NDt-bound} cannot  hold for $t=\max T_s$ if
\[
|T_{s,D}| > \max\left\{\frac{8}{\pmse}\log\left(\frac{TS}{\delta \pmse}\right), \frac{8 C_{t,\delta}}{\pmse \Delta^2}  \right\}
\]
Therefore, 
\[
|T_s| \le |T_{s,D}| + D \le D + \frac{8}{\pmse}\left(\log\left(\frac{TS}{\delta \pmse}\right) + \frac{C_{t,\delta}}{\Delta^2}\right)
\]

Putting everything together, and letting $\Delta=\eps/2$, we obtain that with probability at least $1-3\delta$, the number of times the algorithm chooses $\eps$-bad arms in $\cT$ is bounded by
\begin{align}
\label{eq:eps_freq}
O\left(\frac{T}{W} \frac{SK \log(KWT/\delta)}{\eps^2} + SD + \sum_{s \in [S]}
\frac{1}{\pmse} \frac{\log(TS/\delta)(1-\log(\pmse))}{\eps^2} \right). 
\end{align}

Using the standard peeling technique on $\eps$, we can get an $O((T/W)\log(T)/\Delta_{\min})$ regret bound. First, we can lower bound $\pmse$ by $p_{\min}$, for simplicity. Second, assume that the number of $\eps$-bad action choices is at most $a/\eps^2+b$. Since actions with gap in $\Delta_{\min} [2^k,2^{k+1})$ for $k=0,1,\ldots,\lceil\log_2(1/\Delta_{\min})-1\rceil$ suffer at most $2^{k+1} \Delta_{\min}$ per-round regret,  the total regret is bounded by
\[
\sum_{k=0}^{\lceil\log_2(1/\Delta_{\min})-1 \rceil} 2^{k+1}\Delta_{\min} \left(  \frac{a}{\Delta_{\min}^2} +  b\right)
= \sum_{k=0}^{\lceil\log_2(1/\Delta_{\min})-1 \rceil} \frac{2a}{\Delta_{\min} 2^k} + 2^{k+1}\Delta_{\min} b
\le \frac{4a}{\Delta_{\min}} + 4b.
\]
Applying this to our bound in \eqref{eq:eps_freq} and taking into account that in the windows following the change points, an upper bound on the regret is $W$, we obtain the desired regret bound.


\section{Additional Experimental Results}
\label{ap:figures}

\paragraph{Comparison to the benchmarks in linear scale.}
Here we provide a version of Figure~\ref{fig:all_policies} with a linear scale on the y-axis. This representation helps in identifying that the regret $\UCB$ and $\SWUCB$ grow linearly.

 \begin{figure*}[!th]
     \centering
     \includegraphics[height=0.195\textwidth]{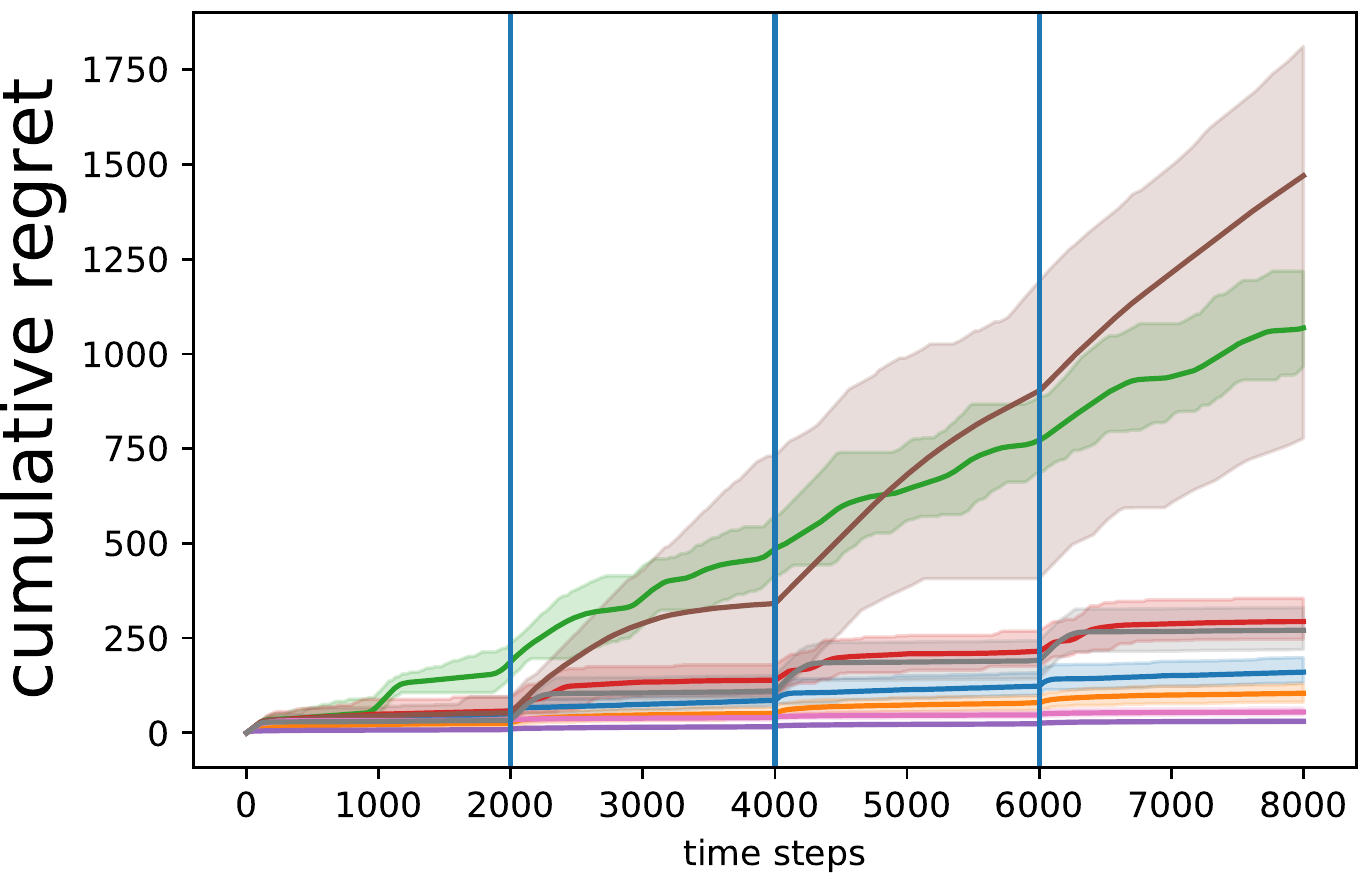}
     \includegraphics[height=0.195\textwidth]{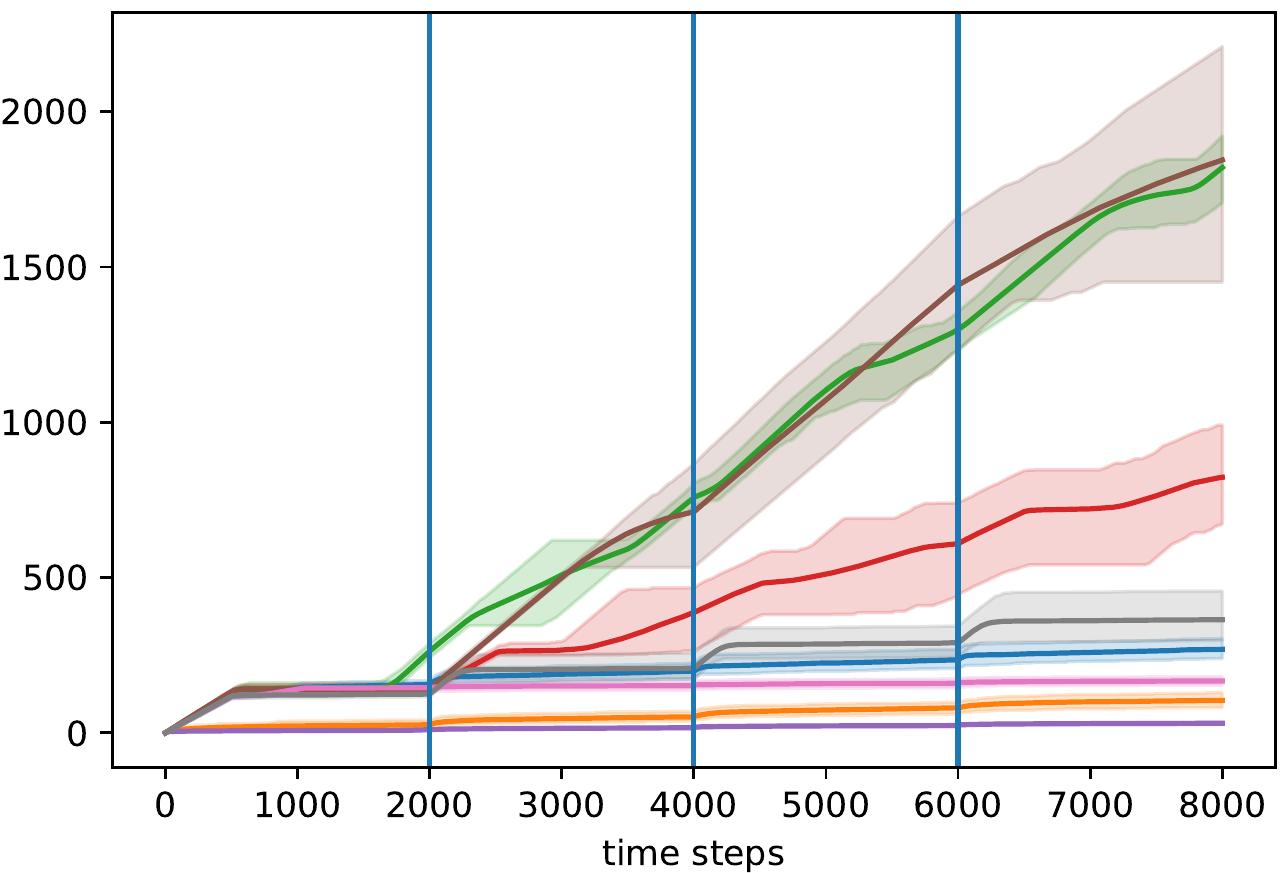}
     \includegraphics[height=0.195\textwidth]{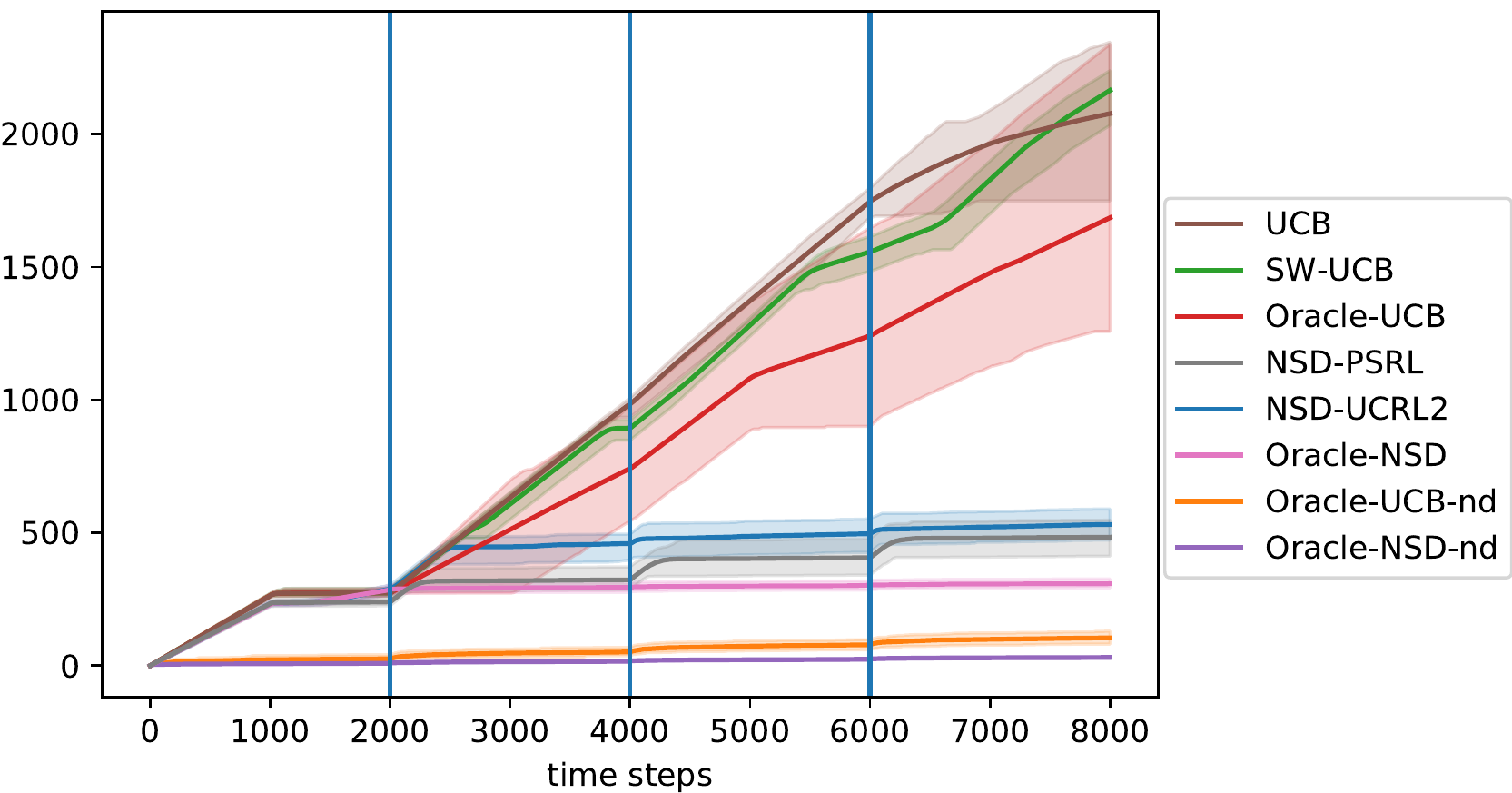}
       \vspace{-0.4cm}
     \caption{Cumulative regret of all policies in linear. $\Gamma_T=3$ and $D\in \{100, 500, 1000\}$ from left to right. When a policy uses a window, $W=800$. Oracles know the change points, non-delayed oracles (nd) receive rewards without delays.  }
     \label{fig:all_policies_linear}
 \end{figure*}

\begin{figure}[!hb]
    \centering
    \includegraphics[width=0.32\textwidth]{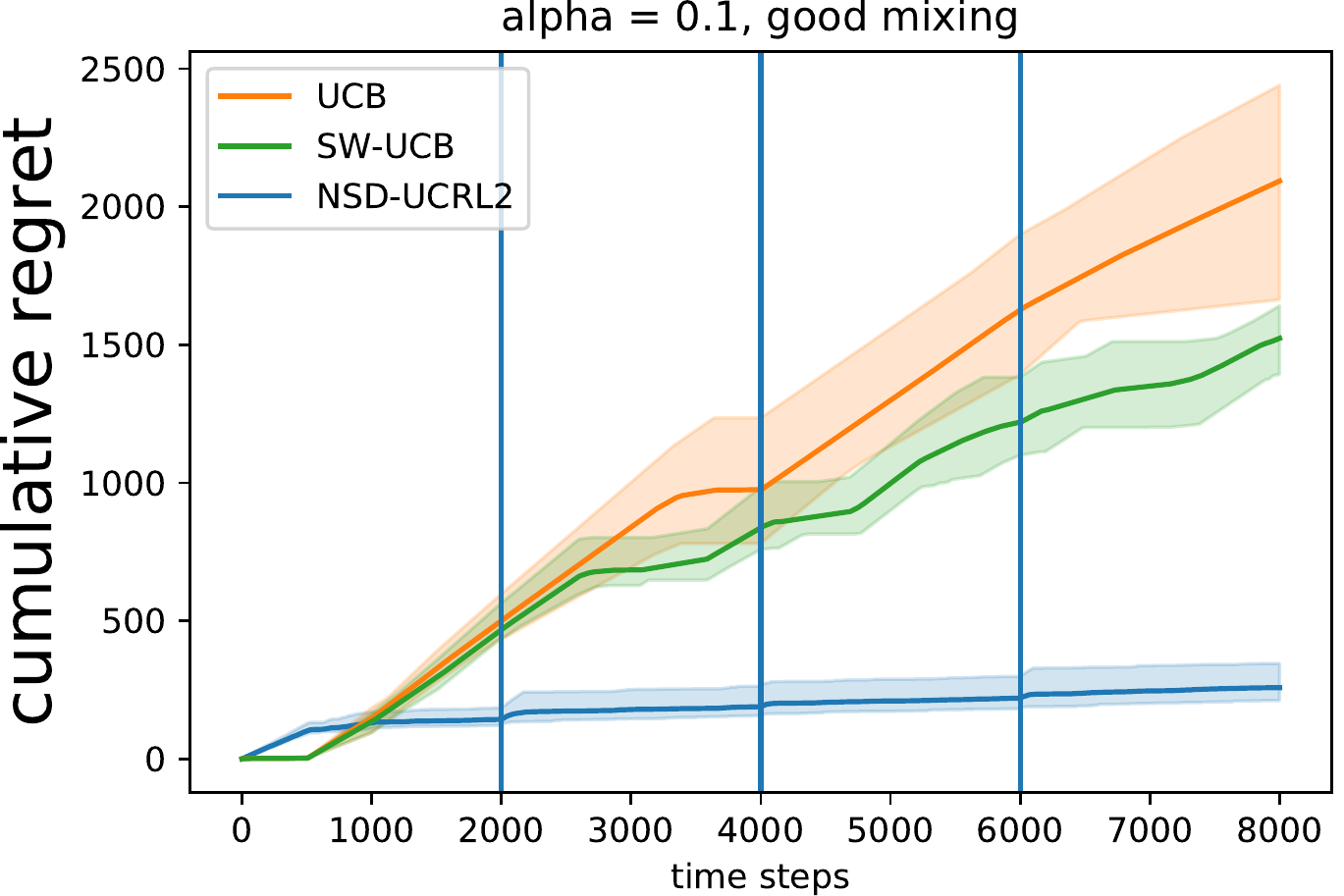}
    \includegraphics[width=0.32\textwidth]{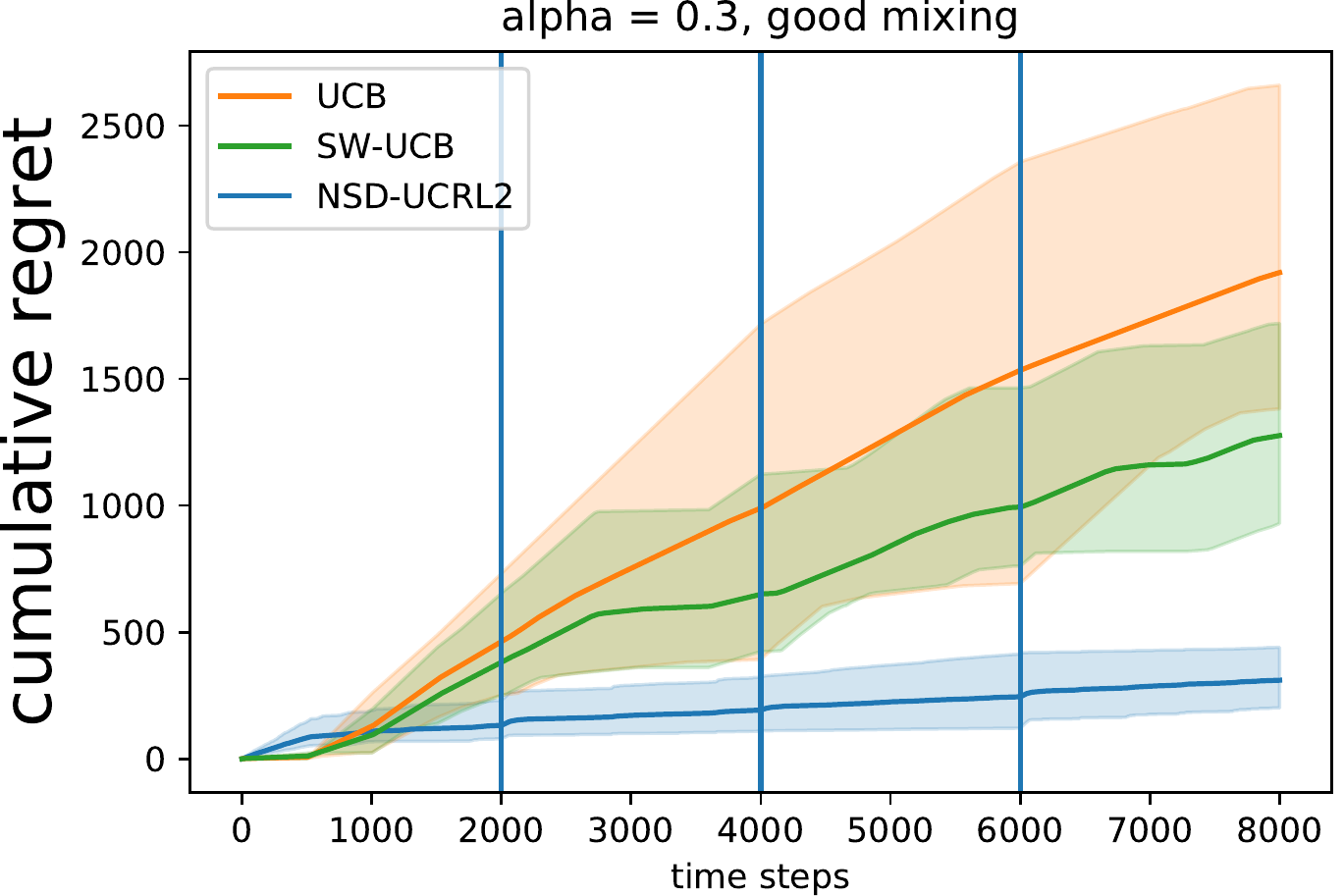}
    \includegraphics[width=0.32\textwidth]{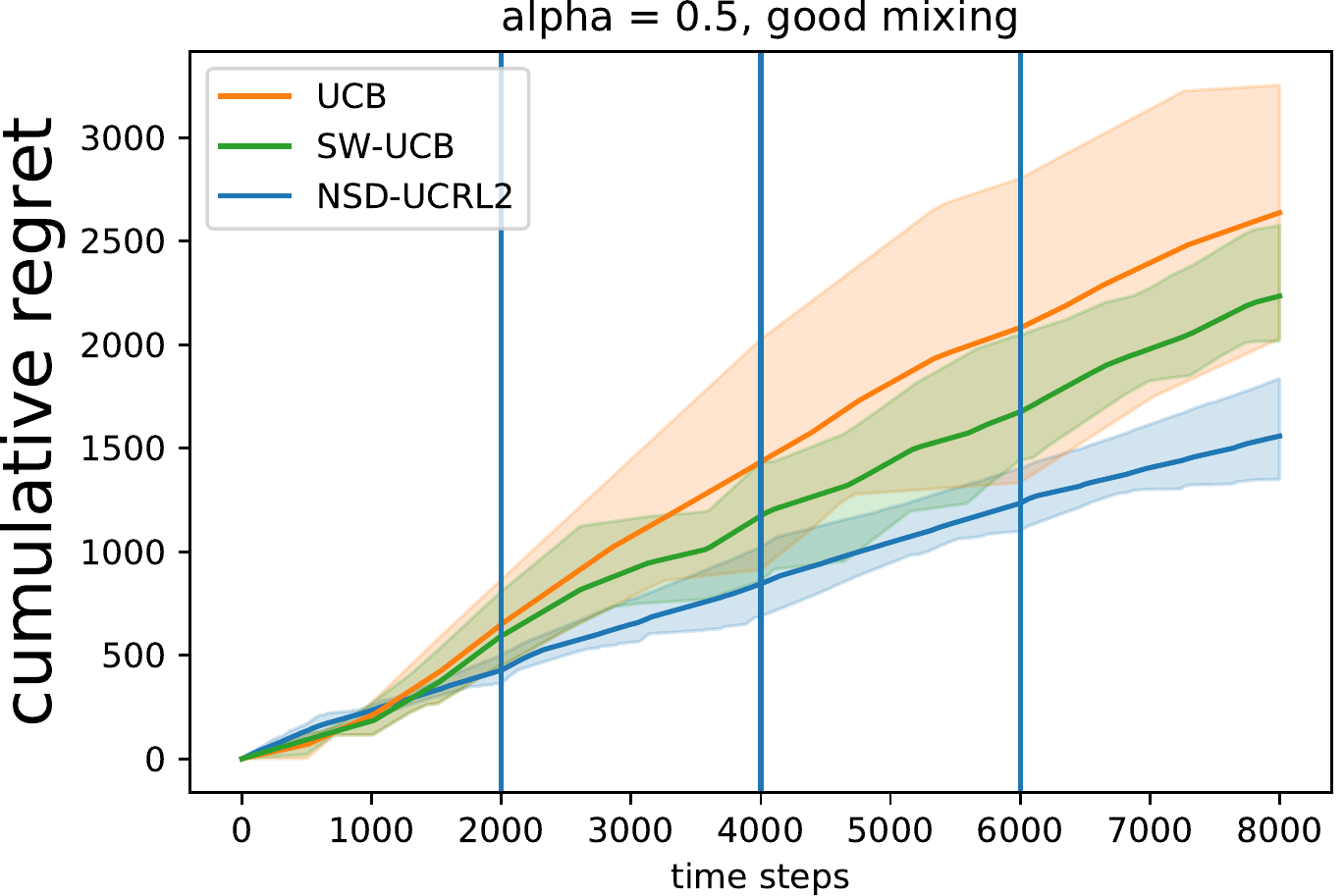}
    \caption{Regret of $\NSDUCRL$, $\SWUCB$ and $\UCB$ in the same setting as Figure~\ref{fig:all_policies} when the model is misspecified. Delays are fixed to $D=500$. From left to right, $\alpha=(0.1, 0.3, 0.5)$ and $\mu=(0.9,0.1,0.1,0.1)$ is chosen so that the best arm changes in the mixed model. Despite the mixing, even for non-negligible mixing levels, $\NSDUCRL$ learns decently fast.}
    \label{fig:nsd_mixed_good}
\end{figure}

\paragraph{NSD bandits with misspecified model. } 
To complement Figure~\ref{fig:nsd_mixed}, we provide in Figure~\ref{fig:nsd_mixed_good} the comparison of $\NSDUCRL$, $\SWUCB$ and $\UCB$ in the favorable case where the best arm is preserved despite the mixing. 

\section{A Bayesian Alternative: $\NSDPSRL$}

\label{ap:psrl}

Bayesian algorithms are popular alternatives to optimistic ones both in bandits \cite{chapelle2011empirical, russo2018tutorial} and in reinforcement learning \cite{osband2017posterior}. We describe a heuristic variant of $\NSDUCRL$ inspired by posterior sampling. This algorithm relies on the prior knowledge of the noise model. In particular, we assume here that the final, delayed rewards are Bernoulli distributed. A similar approach could be followed for Gaussian rewards and other distributions admitting a handy conjugate prior \cite{korda2013thompson}.

The standard approach consists in maintaining posteriors for all parameters of the problem. For each action $a\in [K]$, keep counts $N_t(a,s)$ of every transition to each of the signals $s$ and maintain a Dirichlet posterior $\pi^D_t(a)$ with parameters $(1+N_t(a,s))_s$ for transition probabilities $p(\cdot|a)$. For  each $s\in[S]$, maintain a Beta posterior $\pi^B_t(s)= \text{Beta}(1+\sum_{u=1}^{t-1}R_u(s), 1+N_t(s)-\sum_{u=1}^{t-1}R_t(u))$ for value parameter estimation.
Then, at round $t>0$, all the parameters are sampled from the current posteriors $\tilde{p}_t(a) \sim \pi^D_t(a)$ and $\tilde{\theta}_t(s)\sim \pi^B_t(s)$. 
The value of each action is then computed using directly these samples: $\tilde{\rho}_t(a)= \tilde{p}_t(\cdot|a)^\top \tilde{\theta}_t$, and the algorithm chooses $A_t \in \argmax_{a\in [K]} \tilde{\rho}_t(a)$.

One way to properly extend Thompson sampling to our situation with non-stationary transitions and delayed rewards would be to introduce a prior over the switch distributions for the environment. Instead, in this paper we take the more heuristic approach and adapt the standard method with the sliding-window counts and delayed estimators already defined for $\NSDUCRL$:  the posterior for the transition probability of action $a$ is $\pi^{D,W}_t(a)=\text{Dirichlet}(1+N^W_t(a,1),\ldots,1+N^W_t(a,S))$ and the Beta posterior for $\theta_s$ is computed with the available rewards: $\pi^{B,D}_t(s) = \text{Beta}(1+\sum_{u=1}^{t-D}R_u(s),1+N^D_t(s)-\sum_{u=1}^{t-D}R_u(s))$. $\NSDPSRL$ then selects the best action in this sampled MDP.

%% file: icml_main.bbl
\begin{thebibliography}{32}
\providecommand{\natexlab}[1]{#1}
\providecommand{\url}[1]{\texttt{#1}}
\expandafter\ifx\csname urlstyle\endcsname\relax
  \providecommand{\doi}[1]{doi: #1}\else
  \providecommand{\doi}{doi: \begingroup \urlstyle{rm}\Url}\fi

\bibitem[Agarwal et~al.(2009{\natexlab{a}})Agarwal, Chen, and
  Elango]{agarwal2009spatio}
Agarwal, D., Chen, B.-C., and Elango, P.
\newblock Spatio-temporal models for estimating click-through rate.
\newblock In \emph{Proceedings of the 18th International Conference on World
  Wide Web}, pp.\  21--30. ACM, 2009{\natexlab{a}}.

\bibitem[Agarwal et~al.(2009{\natexlab{b}})Agarwal, Chen, Elango, Motgi, Park,
  Ramakrishnan, Roy, and Zachariah]{agarwal2009online}
Agarwal, D., Chen, B.-C., Elango, P., Motgi, N., Park, S.-T., Ramakrishnan, R.,
  Roy, S., and Zachariah, J.
\newblock Online models for content optimization.
\newblock In \emph{Advances in Neural Information Processing Systems}, pp.\
  17--24, 2009{\natexlab{b}}.

\bibitem[Auer et~al.(2002{\natexlab{a}})Auer, Cesa-Bianchi, and
  Fischer]{auer2002finite}
Auer, P., Cesa-Bianchi, N., and Fischer, P.
\newblock Finite-time analysis of the multiarmed bandit problem.
\newblock \emph{Machine Learning}, 47\penalty0 (2-3):\penalty0 235--256,
  2002{\natexlab{a}}.

\bibitem[Auer et~al.(2002{\natexlab{b}})Auer, Cesa-Bianchi, Freund, and
  Schapire]{auer2002nonstochastic}
Auer, P., Cesa-Bianchi, N., Freund, Y., and Schapire, R.~E.
\newblock The nonstochastic multiarmed bandit problem.
\newblock \emph{SIAM Journal on Computing}, 32\penalty0 (1):\penalty0 48--77,
  2002{\natexlab{b}}.

\bibitem[Auer et~al.(2019)Auer, Chen, Gajane, Lee, Luo, Ortner, and
  Wei]{auer2019achieving}
Auer, P., Chen, Y., Gajane, P., Lee, C.-W., Luo, H., Ortner, R., and Wei, C.-Y.
\newblock Achieving optimal dynamic regret for non-stationary bandits without
  prior information.
\newblock In \emph{Proceedings of the 32nd Conference on Learning Theory}, pp.\
   159--163, 2019.

\bibitem[Azuma(1967)]{azuma1967weighted}
Azuma, K.
\newblock Weighted sums of certain dependent random variables.
\newblock \emph{Tohoku Mathematical Journal, Second Series}, 19\penalty0
  (3):\penalty0 357--367, 1967.

\bibitem[Besbes et~al.(2014)Besbes, Gur, and Zeevi]{besbes2014stochastic}
Besbes, O., Gur, Y., and Zeevi, A.
\newblock Stochastic multi-armed-bandit problem with non-stationary rewards.
\newblock In \emph{Advances in Neural Information Processing Systems}, pp.\
  199--207, 2014.

\bibitem[Cesa-Bianchi \& Lugosi(2012)Cesa-Bianchi and
  Lugosi]{cesa2012combinatorial}
Cesa-Bianchi, N. and Lugosi, G.
\newblock Combinatorial bandits.
\newblock \emph{Journal of Computer and System Sciences}, 78\penalty0
  (5):\penalty0 1404--1422, 2012.

\bibitem[Cesa{-}Bianchi et~al.(2019)Cesa{-}Bianchi, Gentile, and
  Mansour]{CBGM19}
Cesa{-}Bianchi, N., Gentile, C., and Mansour, Y.
\newblock Delay and cooperation in nonstochastic bandits.
\newblock \emph{Journal of Machine Learning Research}, 20\penalty0
  (17):\penalty0 1--38, 2019.

\bibitem[Chapelle(2014)]{chapelle2014modeling}
Chapelle, O.
\newblock Modeling delayed feedback in display advertising.
\newblock In \emph{Proceedings of the 20th ACM SIGKDD International Conference
  on Knowledge Discovery and Data Mining}, pp.\  1097--1105. ACM, 2014.

\bibitem[Chapelle \& Li(2011)Chapelle and Li]{chapelle2011empirical}
Chapelle, O. and Li, L.
\newblock An empirical evaluation of thompson sampling.
\newblock In \emph{Advances in Neural Information Processing Systems}, pp.\
  2249--2257, 2011.

\bibitem[Chen et~al.(2019)Chen, Lee, Luo, and Wei]{chen2019new}
Chen, Y., Lee, C.-W., Luo, H., and Wei, C.-Y.
\newblock A new algorithm for non-stationary contextual bandits: Efficient,
  optimal, and parameter-free.
\newblock In \emph{Proceedings of the 32nd Conference on Learning Theory}, pp.\
   696--726, 2019.

\bibitem[Combes et~al.(2015)Combes, Shahi, Proutiere,
  et~al.]{combes2015combinatorial}
Combes, R., Shahi, M. S. T.~M., Proutiere, A., et~al.
\newblock Combinatorial bandits revisited.
\newblock In \emph{Advances in Neural Information Processing Systems}, pp.\
  2116--2124, 2015.

\bibitem[Dann et~al.(2017)Dann, Lattimore, and Brunskill]{DannLB17}
Dann, C., Lattimore, T., and Brunskill, E.
\newblock Unifying {PAC} and regret: Uniform {PAC} bounds for episodic
  reinforcement learning.
\newblock In \emph{Advances in Neural Information Processing Systems}, pp.\
  5713--5723, 2017.

\bibitem[Gajane et~al.(2018)Gajane, Ortner, and Auer]{gajane2018sliding}
Gajane, P., Ortner, R., and Auer, P.
\newblock A sliding-window algorithm for markov decision processes with
  arbitrarily changing rewards and transitions.
\newblock \emph{arXiv preprint arXiv:1805.10066}, 2018.

\bibitem[Garivier \& Moulines(2011)Garivier and Moulines]{garivier2011upper}
Garivier, A. and Moulines, E.
\newblock On upper-confidence bound policies for switching bandit problems.
\newblock In \emph{Proceedings of the 22nd International Conference on
  Algorithmic Learning Theory}, pp.\  174--188, 2011.

\bibitem[Graves \& Lai(1997)Graves and Lai]{graves1997asymptotically}
Graves, T.~L. and Lai, T.~L.
\newblock Asymptotically efficient adaptive choice of control laws incontrolled
  markov chains.
\newblock \emph{SIAM Journal on Control and Optimization}, 35\penalty0
  (3):\penalty0 715--743, 1997.

\bibitem[Gy{\"{o}}rgy et~al.(2012)Gy{\"{o}}rgy, Linder, and Lugosi]{GyLiLu12}
Gy{\"{o}}rgy, A., Linder, T., and Lugosi, G.
\newblock Efficient tracking of large classes of experts.
\newblock \emph{{IEEE} Transactions on Information Theory}, 58\penalty0
  (11):\penalty0 6709--6725, 2012.

\bibitem[Jaksch et~al.(2010)Jaksch, Ortner, and Auer]{jaksch2010near}
Jaksch, T., Ortner, R., and Auer, P.
\newblock Near-optimal regret bounds for reinforcement learning.
\newblock \emph{Journal of Machine Learning Research}, 11\penalty0
  (51):\penalty0 1563--1600, 2010.

\bibitem[Joulani et~al.(2013)Joulani, Gy\"orgy, and
  Szepesv{\'a}ri]{joulani2013online}
Joulani, P., Gy\"orgy, A., and Szepesv{\'a}ri, {\mbox{Cs}}.
\newblock Online learning under delayed feedback.
\newblock In \emph{Proceedings of the 30th International Conference on Machine
  Learning}, pp.\  1453--1461, 2013.

\bibitem[Korda et~al.(2013)Korda, Kaufmann, and Munos]{korda2013thompson}
Korda, N., Kaufmann, E., and Munos, R.
\newblock Thompson sampling for 1-dimensional exponential family bandits.
\newblock In \emph{Advances in Neural Information Processing Systems}, pp.\
  1448--1456, 2013.

\bibitem[Kveton et~al.(2015)Kveton, Wen, Ashkan, and
  Szepesv\'ari]{kveton2015tight}
Kveton, B., Wen, Z., Ashkan, A., and Szepesv\'ari, {\mbox{Cs}}.
\newblock Tight regret bounds for stochastic combinatorial semi-bandits.
\newblock In \emph{Proceedings of the 18th International Conference on
  Artificial Intelligence and Statistics}, pp.\  535--543, 2015.

\bibitem[Lattimore \& Szepesv{\'a}ri(2020)Lattimore and
  Szepesv{\'a}ri]{lattimore2020bandit}
Lattimore, T. and Szepesv{\'a}ri, {\mbox{Cs}}.
\newblock \emph{Bandit Algorithms}.
\newblock Cambridge University Press, 2020.

\bibitem[Mandel et~al.(2015)Mandel, Liu, Brunskill, and
  Popovi{\'c}]{mandel2015queue}
Mandel, T., Liu, Y.-E., Brunskill, E., and Popovi{\'c}, Z.
\newblock The queue method: Handling delay, heuristics, prior data, and
  evaluation in bandits.
\newblock In \emph{Proceedings of the 29th AAAI Conference on Artificial
  Intelligence}, 2015.

\bibitem[Mann et~al.(2019)Mann, Gowal, Gy{\"{o}}rgy, Hu, Jiang,
  Lakshminarayanan, and Srinivasan]{mann2018learning}
Mann, T.~A., Gowal, S., Gy{\"{o}}rgy, A., Hu, H., Jiang, R., Lakshminarayanan,
  B., and Srinivasan, P.
\newblock Learning from delayed outcomes via proxies with applications to
  recommender systems.
\newblock In \emph{Proceedings of the 36th International Conference on Machine
  Learning}, pp.\  4324--4332, 2019.

\bibitem[Osband \& Van~Roy(2017)Osband and Van~Roy]{osband2017posterior}
Osband, I. and Van~Roy, B.
\newblock Why is posterior sampling better than optimism for reinforcement
  learning?
\newblock In \emph{Proceedings of the 34th International Conference on Machine
  Learning}, pp.\  2701--2710, 2017.

\bibitem[Russo et~al.(2018)Russo, Van~Roy, Kazerouni, Osband, Wen,
  et~al.]{russo2018tutorial}
Russo, D.~J., Van~Roy, B., Kazerouni, A., Osband, I., Wen, Z., et~al.
\newblock A tutorial on thompson sampling.
\newblock \emph{Foundations and Trends{\textregistered} in Machine Learning},
  11\penalty0 (1):\penalty0 1--96, 2018.

\bibitem[Talebi et~al.(2017)Talebi, Zou, Combes, Proutiere, and
  Johansson]{talebi2017stochastic}
Talebi, M.~S., Zou, Z., Combes, R., Proutiere, A., and Johansson, M.
\newblock Stochastic online shortest path routing: The value of feedback.
\newblock \emph{IEEE Transactions on Automatic Control}, 63\penalty0
  (4):\penalty0 915--930, 2017.

\bibitem[Vernade et~al.(2017)Vernade, Capp{\'e}, and
  Perchet]{vernade2017stochastic}
Vernade, C., Capp{\'e}, O., and Perchet, V.
\newblock Stochastic bandit models for delayed conversions.
\newblock In \emph{Proceedings of the 33rd Conference on Uncertainty in
  Artificial Intelligence}, 2017.

\bibitem[Weissman et~al.(2003)Weissman, Ordentlich, Seroussi, Verdu, and
  Weinberger]{weissman2003inequalities}
Weissman, T., Ordentlich, E., Seroussi, G., Verdu, S., and Weinberger, M.~J.
\newblock Inequalities for the l1 deviation of the empirical distribution.
\newblock Technical Report HPL-2003-97R1, Hewlett-Packard Labs., 2003.

\bibitem[Wu et~al.(2017)Wu, Wang, Hong, and Shi]{wu2017returning}
Wu, Q., Wang, H., Hong, L., and Shi, Y.
\newblock Returning is believing: Optimizing long-term user engagement in
  recommender systems.
\newblock In \emph{Proceedings of the 2017 ACM Conference on Information and
  Knowledge Management}, pp.\  1927--1936. ACM, 2017.

\bibitem[Yoshikawa \& Imai(2018)Yoshikawa and Imai]{yoshikawa2018nonparametric}
Yoshikawa, Y. and Imai, Y.
\newblock A nonparametric delayed feedback model for conversion rate
  prediction.
\newblock \emph{arXiv preprint arXiv:1802.00255}, 2018.

\end{thebibliography}
